\documentclass[journal]{IEEEtran}
\pdfoutput=1
\usepackage{xcolor,soul,framed} %,caption

\colorlet{shadecolor}{yellow}
\usepackage[pdftex]{graphicx}
\graphicspath{{../pdf/}{../jpeg/}}
\DeclareGraphicsExtensions{.pdf,.jpeg,.png}
\usepackage{tabularx,booktabs}
\usepackage{lipsum}
\usepackage{float}
\usepackage{algorithm}
\usepackage{array}
\usepackage{mdwmath}
\usepackage{mdwtab}
\usepackage{eqparbox}
\usepackage{url}
\usepackage{amsmath}
\usepackage{algorithmic}
\usepackage{amssymb}
\usepackage{multirow}
\usepackage{verbatim}

\begin{document}

\title{Deep Learning for Image and Point Cloud Fusion in Autonomous Driving: A Review}

\author{Yaodong Cui, ~\IEEEmembership{Student Member,~IEEE,}
Ren Chen, 
Wenbo Chu, 
Long Chen, ~\IEEEmembership{Senior Member,~IEEE,}
Daxin Tian, ~\IEEEmembership{Senior Member,~IEEE,} 
Ying Li,
Dongpu Cao, ~\IEEEmembership{Member,~IEEE}

\thanks{Y. Cui, R. Chen, Y. Li, D. Cao are with the Waterloo CogDrive Lab, Department of Mechanical and Mechatronics Engineering, University of Waterloo, 200 University Ave West, Waterloo ON, N2L3G1 Canada.  e-mail: yaodong.cui@uwaterloo.ca, andrews.rchen@gmail.com, y2424li@uwaterloo.ca, dongpu@uwaterloo.ca (Corresponding author: D. Cao.)}% <-this % stops a space
\thanks{W. Chu is with the China Intelligent and Connected Vehicles (Beijing) Research Institute Co., Ltd., Beijing, 100176, China. e-mail: chuwenbo@china-icv.cn. }
\thanks{L. Chen is  with School of Data and Computer Science, Sun Yat-sen University, Zhuhai 519082, China, and also with Waytous Inc., Qingdao 266109, China. e-mail: chenl46@mail.sysu.edu.cn}% <-this % stops a space
\thanks{D. Tian is with Beijing Advanced Innovation Center for Big Data and Brain Computing, Beijing Key Laboratory for Cooperative Vehicle Infrastructure Systems and Safety Control, School of Transportation Science and Engineering, Beihang University, Beijing 100191, China. }
}

% \markboth{IEEE Transactions on Image Processing,~Vol.~XX, No.~XX, XXX~XXX}
\markboth{Journal of \LaTeX\ Class Files}%
{}
%{Shell \MakeLowercase{\textit{et al.}}: Bare Demo of IEEEtran.cls for Journals}

%  Finally, we provide our insights, observations and points out future research directions.

\maketitle

\begin{abstract}
Autonomous vehicles were experiencing rapid development in the past few years. However, achieving full autonomy is not a trivial task, due to the nature of the complex and dynamic driving environment. Therefore, autonomous vehicles are equipped with a suite of different sensors to ensure robust, accurate environmental perception. In particular, the camera-LiDAR fusion is becoming an emerging research theme. However, so far there has been no critical review that focuses on deep-learning-based camera-LiDAR fusion methods. To bridge this gap and motivate future research, this paper devotes to review recent deep-learning-based data fusion approaches that leverage both image and point cloud. This review gives a brief overview of deep learning on image and point cloud data processing. Followed by in-depth reviews of camera-LiDAR fusion methods in depth completion, object detection, semantic segmentation, tracking and online cross-sensor calibration, which are organized based on their respective fusion levels. Furthermore, we compare these methods on publicly available datasets. Finally, we identified gaps and over-looked challenges between current academic researches and real-world applications. Based on these observations, we provide our insights and point out promising research directions.
\end{abstract}


\IEEEpeerreviewmaketitle

\section{Introduction}
\par Recent breakthroughs in deep learning and sensor technologies have motivated rapid development of autonomous driving technology, which could potentially improve road safety, traffic efficiency and personal mobility  \cite{Duarteeaav9843} \cite{8760560} \cite{bigman2020life}. However, technical challenges and the cost of exteroceptive sensors have constrained current applications of autonomous driving systems to confined and controlled environments in small quantities. One critical challenge is to obtain an adequately accurate understanding of the vehicle’s 3D surrounding environment in real-time. To this end, sensor fusion, which leverages multiple types of sensors with complementary characteristics to enhance perception and reduce cost, has become an emerging research theme.

\begin{figure}
\centering
\includegraphics[scale=0.18]{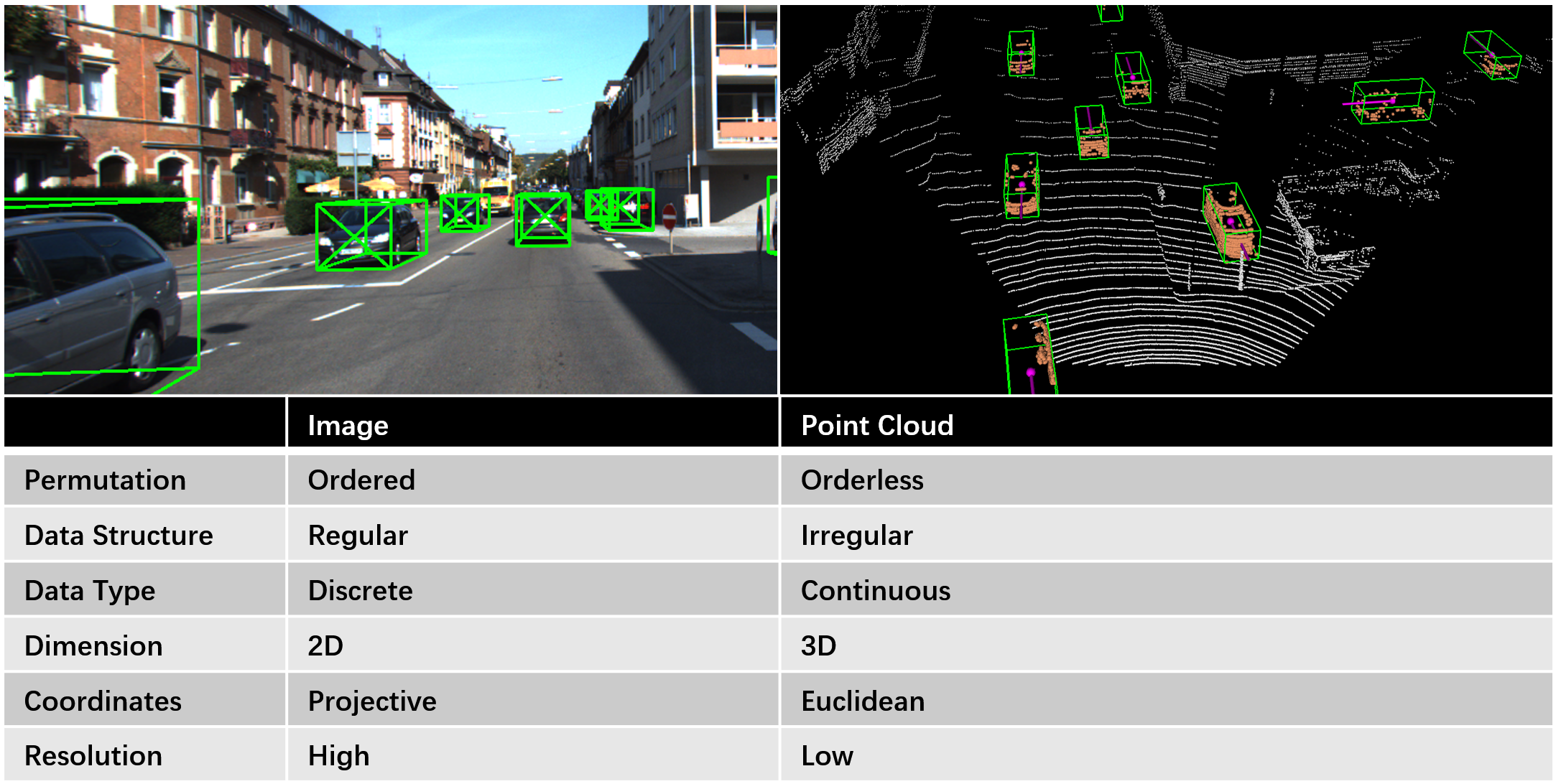}
\caption{ A comparison between image data and point cloud data. }
\label{comp1}
\end{figure}

\begin{figure*}
\centering
\includegraphics[scale=0.48]{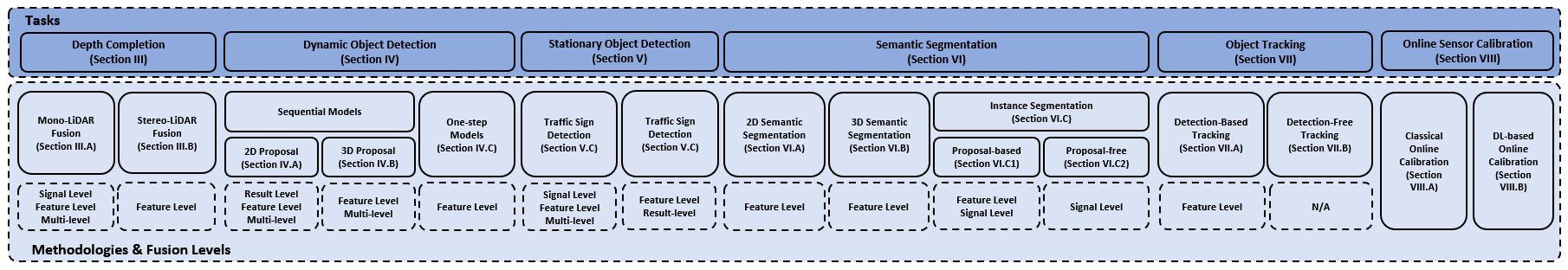}
\caption{Tasks related to image and point cloud fusion based perception and their corresponding sections.}
\label{taxonomy}
\end{figure*}

\par In particular, recent deep learning advances have significantly improved the performance of camera-LiDAR fusion algorithms. Cameras and LiDARs have complementary characteristics, which make camera-LiDAR fusion models more effective and popular compared with other sensor fusion configurations (radar-camera, LiDAR-radar, etc.,). To be more specific, vision-based perception systems achieve satisfactory performance at low-cost, often outperforming human experts \cite{Silver2016MasteringTG} \cite{mnih2015humanlevel}. However, a mono-camera perception system cannot provide reliable 3D geometry, which is essential for autonomous driving \cite{7819480} \cite{8886732}. On the other hand, stereo cameras can provide 3D geometry, but do so at high computational cost and struggle in high-occlusion and textureless environments \cite{7322252} \cite{8624583} \cite{8701620}. Furthermore, camera base perception systems struggle with complex or poor lighting conditions, which limit their all-weather capabilities \cite{8419782}. Contrarily, LiDAR can provide high-precision 3D geometry and is invariant to ambient light. However, mobile LiDARs are limited by low-resolution (ranging from 16 to 128 channels), low-refresh rates (10Hz), severe weather conditions (heavy rain, fog and snow) and high cost. To mitigate these challenges, many works combined these two complementary sensors and demonstrated significant performance advantages than a-modal approaches. Therefore, this paper focuses on reviewing current deep learning fusion strategies for camera-LiDAR fusion. 

% Therefore, by combining image and point cloud, we can make the best of both sensory data. images record the real-world via projection to the 
\par Camera-LiDAR fusion is not a trivial task. First of all, cameras record the real-world by projecting it to the image plane, whereas the point cloud preserves the 3D geometry. Furthermore, in terms of data structure, the point cloud is irregular, orderless and continuous, while the image is regular, ordered and discrete. These characteristics differences between the point cloud and the image lead to different feature extraction methodologies. In Figure 1, a comparison between the characteristics of image and point are shown.

% To the best of our knowledge, this paper is the first review that focuses on deep learning approaches of camera-LiDAR fusion.
\par  Previous reviews \cite{wang2019multi} \cite{Feng_2020} on deep learning methods for multi-modal data fusion covered a broad range of sensors, including radars, cameras, LiDARs, Ultrasonics, IMU, Odometers, the GNSS and HD Maps. This paper focuses on camera-LiDAR fusion only and therefore is able to present more detailed reviews on individual methods. Furthermore, we cover a broader range of perception related topics (depth completion, dynamic and stationary object detection, semantic segmentation, tracking and online cross-sensor calibration) that are interconnected and are not fully included in the previous reviews \cite{Feng_2020}. The contribution of this paper is summarized as the following:
\begin{itemize}
    \item To the best of our knowledge, this paper is the first survey focusing on deep learning based image and point cloud fusion approaches in autonomous driving, including depth completion, dynamic and stationary object detection, semantic segmentation, tracking and online cross-sensor calibration.
    \item This paper organizes and reviews methods based on their fusion methodologies. Furthermore, this paper presented the most up-to-date (2014-2020) overviews and performance comparisons of the state-of-the-art camera-LiDAR fusion methods.
    \item This paper raises overlooked open questions, such as open-set detection and sensor-agnostic framework, that are critical for the real-world deployment of autonomous driving technology. Moreover, summaries of trends and possible research directions on open challenges are presented.
\end{itemize}
\par This paper first provides a brief overview of deep learning methods on image and point cloud data in Section II. In Sections III to VIII, reviews on camera-LiDAR based depth completion, dynamic object detection, stationary object detection, semantic segmentation, object tracking and online sensor calibration are presented respectively. Trends, open challenges and promising directions are discussed in Section VII. Finally, a summary is given in Section VIII. Figure 2 presents the overall structures of this survey and the corresponding topics.

\section{A Brief Review of Deep Learning}
% \par This section present a overview of artificial intelligence and an introduction to deep learning on image and point cloud. 
% \subsection{A Brief Overview of Artificial Intelligence}
% \par Deep learning is a subset of artificial neural networks which leverages multiple network layers to extract features progressively. The early works of learning based artificial neural networks start in the 1940s and gained increasing attention and developments throughout the 1970s and 1980s. Some of the most important concepts, such as automatic differentiation \cite{1100330}, backpropagation \cite{linnainmaa1976taylor} \cite{werbos1990backpropagation}, max-pooling \cite{287150} are proposed during this time. During the 1990s and 2000s, researches in artificial intelligence have witnessed a slow down due to various reasons. However, the exponential growth of computational power and data, combined with advances in network architecture and training strategies, researches in artificial intelligence has since regaining momentum. Especially, deep learning based methods have demonstrated great potentials, outperforming the other methods in the ImageNet competition \cite{RussakovskyFeiFei} \cite{NIPS2012_4824} \cite{szegedy2015going}.
% \par 

\subsection{Deep Learning on Image}

\par Convolutional Neural Networks (CNNs) are one of the most efficient and powerful deep learning models for image processing and understanding. Compared to the Multi-Layer-Perceptron (MLP), the CNN is shift-invariant, contains fewer weights and exploits hierarchical patterns, making it highly efficient for image semantic extraction. Hidden layers of a CNN consist of a hierarchy of convolutional layers, batch normalization layers, activation layers and pooling layers, which are trained end-to-end. This hierarchical structure extracts image features with increasing abstract levels and receptive ﬁelds, enabling the learning of high-level semantics.

% \begin{itemize}
% \item Hierarchical feature representation, which is the multilevel representations from pixel to high-level semantic features learned by a hierarchical multi-stage structure [15], [53], can be learned from data automatically and hidden factors of input data can be disentangled through multi-level nonlinear mappings.
% \item Compared with traditional shallow models, a deeper architecture provides an exponentially increased expressive capability. 
% \item The architecture of CNN provides an opportunity to jointly optimize several related tasks together (e.g. Fast RCNN combines classiﬁcation and bounding box regression into a multi-task leaning manner). 
% \item Beneﬁtting from the large learning capacity of deep CNNs, some classical computer vision challenges can be recast as high-dimensional data transform problems and solved from a different viewpoint.
% \end{itemize}

\subsection{Deep Learning on Point Cloud}

\par The point cloud is a set of data points, which are LiDAR's measurements of the detected object's surface. In terms of data structure, the point cloud is sparse, irregular, orderless and continuous. Point cloud encodes information in 3D structures and in per-point features (reflective intensities, color, normal, etc.,), which is invariant to scale, rigid transformation and permutation. These characteristics made feature extractions on the point cloud challenging for existing deep learning models, which require the modifications of existing models or developing new models. Therefore, this section focuses on introducing common methodologies for point cloud processing.

 \subsubsection{Volumetric representation based} 
\par The volumetric representation partitions the point cloud into a fixed-resolution 3D grid, where features of each grid/voxel are hand-crafted or learned. This representation is compatible with standard 3D convolution \cite{Zhou_2018} \cite{7353481} \cite{8941003}. Several techniques have been proposed in \cite{Qi_2016} to reduce over-fittings, orientation sensitivity and capture internal structures of objects. However, the volumetric representation loses spatial resolution and fine-grained 3D geometry during voxelization which limits its performance. Furthermore, attempts to increase its spatial resolution (denser voxels) would cause computation and memory footprint to grow cubically, making it unscalable.

 \subsubsection{Index/Tree representation based} 
\par To alleviate constraints between high spatial resolution and computational costs, adapted-resolution partition methods that leverage tree-like data structures, such as kd-tree \cite{zeng20183dcontextnet} \cite{Klokov_2017}, octree \cite{riegler2017octnet} \cite{8784388} \cite{lei2019octree} are proposed. By dividing the point cloud into a series of unbalanced trees, regions can be partitioned based on their point densities. This allows regions with lower point densities to have lower resolutions, which reduce unnecessary computation and memory footprint. Point features are extracted along with the pre-built tree structure.

 \subsubsection{2D views representation based} 
2D views/multi-views are generated by projecting the point cloud to multiple 2D view planes. These rendered multi-view images can be processed by standard 2D convolutions and features from these views are aggregated via view-pooling layers \cite{su2015multi}.  Thus, the permutation-invariant problem is solved by transforming point cloud to images and the translation-invariant is achieved by aggregating features from different views. Qi et al. \cite{Qi_2016} combined the volumetric representation with multi-views generated via sphere rendering. Unfortunately, 2D views methods lose the 3D geometry information during the view rendering and struggle with per-point label prediction \cite{Klokov_2017}.

\subsubsection{Graph representation based}
Point clouds can be represented as graphs and convolution-like operation can be implemented on graphs in the spatial or spectral domain \cite{8931245} \cite{henaff2015deep} \cite{simonovsky2017dynamic}. For graph-convolution in the spatial domain, operations are carried out by MLPs on spatially neighbouring points. Spectral-domain graph-convolutions extend convolutions as spectral filtering on graphs through the Laplacian Spectrum  \cite{doi:10.1111/cgf.12693} \cite{ae482107de73461787258f805cf8f4ed} \cite{defferrard2016convolutional}.

\subsubsection{Point representation based} 
Point representation based methods consume the point cloud without transforming it into an intermediate data representation. Early works in this direction employ shared Multi-Layer Perceptrons (MLPs) to process point cloud \cite{qi2017pointnet} \cite{Charles_2017} , while recent works concentrated on defining specialized convolution operations for points \cite{hua2018pointwise} \cite{Atzmon_2018} \cite{8943956} \cite{NIPS2018_7362} \cite{Lan_2019} \cite{Groh_2019} \cite{Thomas_2019}.  

\par One of the pioneering works of direct learning on point clouds is the PointNet \cite{Charles_2017} \cite{qi2017pointnet}, which employs an independent T-Net module to align point clouds and shared MLPs to process individual points for per-point feature extraction. The computation complexity of the PointNet increases linearly with the number of inputs, making it more scalable compared with volumetric based methods. To achieve permutation invariant, point-wise features are extracted by shared MLPs which are identical for all points. These features are aggregated by symmetric operations (i.e. max-pooling), which are also permutation invariant. The feature extraction process of the PointNet is defined as:
% The key idea which allows the processing of unordered data, is to employ symmetric functions (e.g. max-pooling):  
\begin{equation}\label{eq1}
g\left(\left\{x_{1}, \ldots, x_{n}\right\}\right) \approx f_{sym}\left(h\left(x_{1}\right), \ldots, h\left(x_{n}\right)\right)
\end{equation}
where $x$ represents input points, $h$ represents the per-point feature extraction function (i.e. shared MLPs), $f_{sym}$ represents a symmetric function (i.e. max-polling) and $g$ is a general function that we want to approximate. 
\par However, the PointNet fails to extract local inter-point geometry at different levels. To mitigate this challenge, Qi et al. \cite{qi2017pointnet} extended the PointNet to extract features from different levels by grouping points into multiple sets and apply PointNets locally. To reduce the computational and memory cost of the PointNet++ \cite{qi2017pointnet}, the RandLA-Net \cite{hu2020randla} stacked the random point sampling modules and attention-based local feature aggregation modules hierarchically to progressively increase receptive field while maintaining high efficiency.

\par Unlike PointNet-based methods, the spatial relationship between points is explicitly modelled in point-wise convolutions. Point-wise convolutions aim to generalize the standard 2D discrete convolution to the continuous 3D space. The main challenge is to replace the discrete weight filter in standard convolution with a continuous weight function. This continuous weight function is approximated using MLPs in PointConv \cite{Wu_2019} and correlation functions in KPConv \cite{Thomas_2019} and PCNN \cite{Atzmon_2018}. More specifically, the PCNN \cite{Atzmon_2018} defines convolution kernels as 3D points with weights. A Gaussian correlation function that takes the coordinates of the kernel point and input point is used to calculate the weighting matrix at any given 3D coordinates. The KPConv \cite{Thomas_2019} follows this idea but instead uses a linear correlation function. Furthermore, KPConvs \cite{Thomas_2019} are applied on local point patches hierarchically, which are similar to the concepts of standard CNNs. This general point-wise convolution $\mathcal{F}$ at an input point $x \in \mathbb{R}^{3}$ in 3D continuous space is defined as:
\begin{equation}\label{eq5}
(\mathcal{F} * h)(x)=\sum_{x_{i} \in \mathcal{N}_{x}} h\left(x_{i}-x\right) f_{i} 
\end{equation}
where $h$ is the per-point kernel function which calculates the weighting matrix given the coordinates of input points and kernel points. $x_{i}$ and $f_{i}$ are the $ith$ neighboring points of $x$ and their corresponding features (intensity, color etc.,). $\mathcal{N}_{x}$ are all the neighboring points of the input point $x$, which are determined using KNN or radius neighborhoods.
% \par PointCNN \cite{NIPS2018_7362} attempts to achieve permutation-invariant by permute points using a learnt transformation matrix ($\mathcal{X}$ -transformations). However, the author found this learnt $\mathcal{X}$ -transformations to be less ideal. 

\begin{table*}[]
\centering
\caption{Comparative results on KITTI depth completion benchmark \cite{Uhrig2017THREEDV}. 'LL','S' , 'SS','U'  stand for learning scheme, supervised, self-supervised and unsupervised respectively. 'RGB','sD','Co', 'S',stand for RGB image data , sparse depth, confidence and stereo disparity. 'iRMSE','iMAE', 'RMSE','MAE' stand for root mean squared error of the inverse depth [1/km], mean absolute error of the inverse depth [1/km], root mean squared error [mm], mean absolute error [mm].}
\label{tab:depth}
\centering
\begin{tabular}{|l|l|l|l|l|l|c|c|c|c|c|c|} 
\cline{1-12}
\multirow{2}{*}{Methods}                                                                          & Fusion                   & \multirow{2}{*}{LL} &\multirow{2}{*}{Input}           & \multirow{2}{*}{Models}     & \multirow{2}{*}{Hardware}         &Run & Model & \multirow{2}{*}{iRMSE} & \multirow{2}{*}{iMAE} & \multirow{2}{*}{RMSE}    &\multirow{2}{*}{MAE}    \\ 
                                                                                                     & Level                 &      &                                 &                             &    &Time & Size &  &  &     &     \\ 
\cline{1-12}
\multirow{8}{*}{\begin{tabular}[c]{@{}l@{}}Mono-\\ LiDAR\\ Fusion \end{tabular}} & \multirow{4}{*}{\begin{tabular}[c]{@{}l@{}}Signal \\ Level \end{tabular}}  & S   &RGB+sD   & Sparse2Dense \cite{Ma_2018}     & Tesla P100  & 0.08s & 12M   & 4.07  & 1.57 & 1299.85 & 350.32  \\
                                                                                 &                                & SS  &RGB+sD    & Sparse2Dense++ \cite{Ma_2019}   & Tesla V100   & 0.08s &22M   & 2.80  & 1.21 & 814.73  & 249.95  \\
                                                                                 &                                & S   &RGB+sD   & CSPN \cite{Cheng_2018}          & Titan X       & 1s     & 25M      & 2.93  & 1.15 & 1019.64 & 279.46  \\
                                                                                 &                                & S   &RGB+sD   & CSPN++ \cite{cheng2019cspn}     & Tesla P40     & 0.2s   & -  & 2.07  & 0.90 & 743.69  & 209.28  \\ 
\cline{2-12}
                                                                                 & \multirow{3}{*}{\begin{tabular}[c]{@{}l@{}}Feature \\ Level \end{tabular}} & S  &RGB+sD    & Spade-RGBsD \cite{Jaritz_2018}     & N/A   & - & -   & 2.17  & 0.95 & 917.64  & 234.81  \\
                                                                                 &                                & S  &RGB+sD+Co & NConv-CNN \cite{Eldesokey_2019}    & Tesla V100    & 0.02s & 0.5M   & 2.60  & 1.03 & 829.98  & 233.26  \\
                                                                                 &                                & S  &RGB+sD   & GuideNet \cite{tang2019learning}    & GTX 1080Ti   & 0.14s & -   & 2.25  & 0.99 & 736.24  & 218.83  \\ 
\cline{2-12}
                                                                                 & Multi                    & U  &RGB+sD  & RGBguide \cite{Van_Gansbeke_2019} & Tesla V100  & 0.02s & 0.35M   & 2.19  & 0.93 & 772.87  & 215.02  \\ 
\cline{1-12}
\multirow{3}{*}{\begin{tabular}[c]{@{}l@{}}Stereo-\\ LiDAR \end{tabular}}        & \multirow{3}{*}{\begin{tabular}[c]{@{}l@{}}Feature \\ Level \end{tabular}} & S  &sD+S    & HDE-Net \cite{8624583}   & Titan X   & 0.05s  & 4.2M  & -     & -    & -       & -       \\
                                                                                 &                                & U  &sD+S    &LidarStereoNet \cite{Cheng_2019}  & Titan X   & -   & -     & -     & -    & -       & -       \\
                                                                                 &                                & SS &sD+S    & LiStereo \cite{zhang2019listereo} &  Titan X  & -       & -        & 2.19  & 1.10 & 832.16  & 283.91  \\
\cline{1-12}
\end{tabular}

\end{table*}

\begin{figure}
\centering
\includegraphics[scale=0.21]{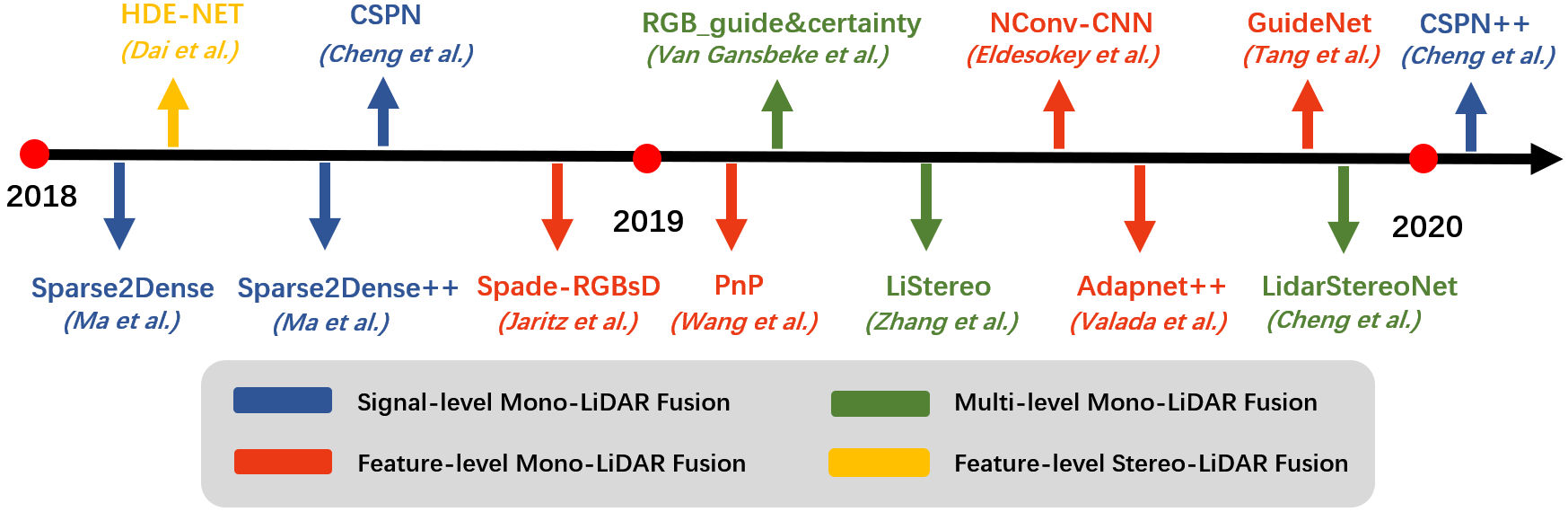}
\caption{ Timeline of depth completion models and their corresponding fusion levels. }
\label{structure}
\end{figure}

\section{Depth Completion}
% \par This section 
\par Depth Completion aims to up-sample sparse irregular depth to dense regular depth, which facilitates the down-stream perception module. Depth completion can reduce the drastic uneven distributions of points in a LiDAR scan. For instance, far-away objects represented by a hand full of points are up-sampled to match their closer counterparts. To achieve this, high-resolution images are often employed to guide the 3D depth up-sampling. The depth completion task can be represented as:
\begin{equation}\label{eq6}
w^{*}=\underset{w}{\arg} \min  \mathcal{L}(f(x ; w), G)
\end{equation}
where the network $f(.)$ parametrized by $w$, predicts the ground truth $G$, given the input $x$. The loss function is represented as $\mathcal{L}(\cdot, \cdot)$.
\par Figure 3 gives the timeline of depth completion models and their corresponding fusion levels. The comparative results of depth completion models on the KITTI depth completion benchmark \cite{Uhrig2017THREEDV} are listed in Table I.

% which can be categorized into signal-level fusion, feature-level fusion and multi-level (signal-level \& feature-level) fusion. Most deep learning based approaches are encoder-decoder/autoencoder based, which originated from the generative model.
% There are several benefits of using stereo image pairs  

\subsection{Mono Camera and LiDAR fusion}
% \par Most current researches on depth completion focused on using images from a mono camera to guide depth completion. 
The idea behind image-guided depth completion is that dense RGB/color information contains relevant 3D geometry. Therefore, images can be leveraged as a reference for depth up-sampling. 
\subsubsection{Signal-level fusion}
\par In 2018, Ma et al. \cite{Ma_2018} presented a ResNet \cite{He_2016} based autoencoder network that leverages RGB-D images (i.e. images concatenated with sparse depth maps) to predict dense depth maps. However, this method requires pixel-level depth ground truth, which is difficult to obtain. To solve this issue, Ma et al. \cite{Ma_2019} presented a model-based self-supervised framework that only requires a sequence of images and sparse depth images for training. This self-supervision is achieved by employing sparse depth constrain, photometric loss and smoothness loss. However, this approach assumes objects to be stationary. Furthermore, the resulting depth output is blurry and input depth may not be preserved. 
\par To generate a sharp dense depth map in real-time, Cheng et al. \cite{Cheng_2018} fed RGB-D images to a convolutional spatial propagation network (CSPN). This CSPN aims to extract the image-dependent affinity matrix directly, producing significantly better results in key measurements with lesser run-time. In CSPN++, Cheng et al. \cite{cheng2019cspn} proposed to dynamically select convolutional kernel sizes and iterations to reduce computation. Furthermore, CSPN++ employs weighted assembling to boost its performance. 

\subsubsection{Feature-level fusion}
Jaritz et al. \cite{Jaritz_2018} presented an autoencoder network that can either perform depth completion or semantic segmentation from sparse depth maps and images without applying validity masks. Images and sparse depth maps are first processed by two parallel NASNet-based encoders \cite{Zoph_2018} before fusing them into the shared decoder. This approach can achieve decent performance with very sparse depth inputs (8-channel LiDAR). Wang et al. \cite{8794404} designed an integrable module (PnP) that leverages the sparse depth map to improve the performance of existing image-based depth prediction networks. This PnP module leverages gradient calculated from sparse depth to update the intermediate feature map produced by the existing depth prediction network.  Eldesokey et al. \cite{Eldesokey_2019} presented a framework for unguided depth completion that processes images and very sparse depth maps in parallel and combine them in a shared decoder. Furthermore, normalized convolutions are used to process highly sparse depth and to propagate confidence. Valada et al. \cite{Valada_2019} extended one-stage feature-level fusion to multiple-stages of varying depth of the network. Similarly, GuideNet \cite{tang2019learning} fuse image features to sparse depth features at different stages of the encoder to guide the up-sampling of sparse depths, which achieves top performance in the KITTI depth completion benchmark. The constrain of these approaches is the lack of large-scale datasets that have dense depth ground truth.

\begin{figure*}
\centering
\includegraphics[scale=0.25]{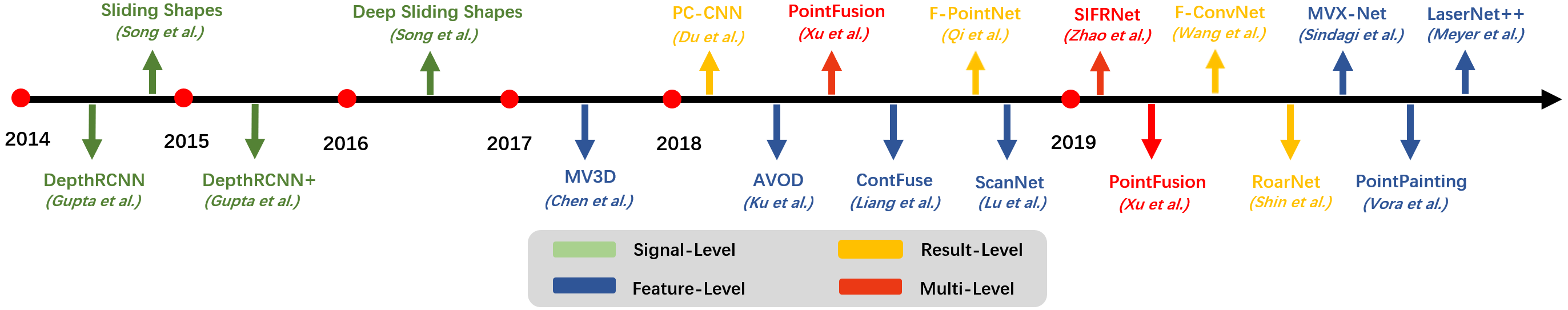}
\caption{ Timeline of 3D object detection networks and their corresponding fusion levels. }
\label{timelineobj}
\end{figure*}

% \begin{figure}
% \centering
% \includegraphics[scale=0.27]{scatter2.png}
% \caption{ A scatter plot of models' root mean squared error of the inverse depth ($iMRSE$) with respect to runtime. The results are based on the public available KITTI test set. }
% \label{scatter2}
% \end{figure}

% \begin{figure}
% \centering
% \includegraphics[scale=0.18]{depth arch.png}
% \caption{Some typical model architectures and fusion methods for depth completion}
% \label{arch1}
% \end{figure}

\subsubsection{Multi-level fusion}
Van Gansbeke et al. \cite{Van_Gansbeke_2019} further combines signal-level fusion and feature-level fusion in an image-guided depth completion network. The network consists of a global and a local branch to process RGB-D data and depth data in parallel before fusing them based on the confidence maps.

\subsection{Stereo Cameras and LiDAR fusion}
\par Compared with the RGB image, dense depth disparity from stereo cameras contains richer ground truth 3D geometry. On the other hand, LiDAR depth is sparse but of higher accuracy. These complementary characteristics enable stereo-LiDAR fusion based depth completion models to produce a more accurate dense depth. However, it is worth noting that stereo cameras have limited range and struggles in high-occlusion, texture-less environments, making them less ideal for autonomous driving.
\subsubsection{Feature-level fusion}
\par  One of the pioneering works is from Park et al. \cite{8624583}, in which high-precision dense disparity map is computed from dense stereo disparity and point cloud using a two-stage CNN. The first stage of the CNN takes LiDAR and stereo disparity to produce a fused disparity. In the second stage, this fused disparity and left RGB image is fused in the feature space to predict the final high-precision disparity. Finally, the 3D scene is reconstructed from this high-precision disparity. The bottleneck of this approach is the lack of large-scale annotated stereo-LiDAR datasets. The LidarStereoNet \cite{Cheng_2019} averted this difficulty with an unsupervised learning scheme, which employs image warping/photometric loss, sparse depth loss, smoothness loss and plane fitting loss for end-to-end training. Furthermore, the introduction of 'feedback loop' makes the LidarStereoNet robust against noisy point cloud and sensor misalignment. Similarly, Zhang et al. \cite{zhang2019listereo} presented a self-supervised scheme for depth completion. The loss function consists of sparse depth, photometric and smoothness loss.

\section{Dynamic Object Detection}
% \subsection{Mono Camera and LiDAR fusion}

\par  Object detection (3D) aims to locate, classify and estimate oriented bounding boxes in the 3D space. This section is devoted to dynamic object detection, which includes common dynamic road objects (car, pedestrian, cyclist, etc.,). There are two main approaches for object detection: sequential and one-step. Sequential based models consist of a proposal stage and a 3D bounding box (bbox) regression stage in the chronological order. In the proposal stage, regions that may contain objects of interest are proposed. In the bbox regression stage, these proposals are classified based on the region-wise features extracted from 3D geometry. However, the performance of sequential fusion is limited by each stage. On the other hand, one-step models consist of one stage, where 2D and 3D data are processed in a parallel manner. 
\par The timeline of 3D object detection networks and typical model architectures are shown in Figures 4 \& 5. Table II presents the comparative results of 3D object detection models on the KITTI 3D Object Detection benchmark \cite{Geiger2012CVPR}. Table III summarizes and compares dynamic object detection models.

\subsection{2D Proposal Based Sequential Models}

\par A 2D proposal based sequential model attempts to utilize 2D image semantics in the proposal stage, which takes advantage of off-the-shelf image processing models. Specifically, these methods leverage the image object detector to generate 2D region proposals, which are projected to the 3D space as detection seeds. There are two projection approaches to translate 2D proposals to 3D. The first one is projecting bounding boxes in the image plane to the point cloud, which results in a frustum shaped 3D search space. The second method projects the point cloud to the image plane, which results in the point cloud with point-wise 2D semantics. 
% However, far away or occluded objects are often represented by a handful of sparse points, making 3D bbox regression difficult.

\subsubsection{Result-level Fusion}
\par The intuition behind result-level fusion is to use off-the-shelf 2D object detectors to limit the 3D search space for 3D object detection, which significantly reduces computation and improve run-time. However, since the whole pipeline depends on the results of the 2D object detector, it suffers from the limitations of the image-based detector. 

% In F-PointNet, the fusion of multimodal data is only used to reduce computation.

\par One of the early works of result-level fusion is the F-PointNets \cite{Qi_2018}, where 2D bounding boxes are first generated from images and projected to the 3D space. The resulting projected frustum proposals are fed into a PointNet\cite{Charles_2017} based detector for 3D object detection.  Du et al. \cite{Du_2018} extended the 2D to 3D proposals generation stage with an additional proposal refinement stage, which further reduces unnecessary computation on the background point. During this refinement stage, a model fitting based method is used to filter out background points inside the seed region. Finally, the filtered points are fed into the bbox regression network. The RoarNet \cite{Shin_2019} followed a similar idea, but instead employs a neural network for the proposal refinement stage. Multiple 3D cylinder proposals are first generated based on each 2D bbox using the geometric agreement search \cite{Mousavian_2017}, which results in smaller but more precise frustum proposals then the F-pointNet \cite{Qi_2018}. These initial cylinder proposals are then processed by a PointNet \cite{qi2017pointnet} based header network for the final refinement. To summarize, these approaches assume each seed region only contains one object of interest, which is however not true for crowded scenes and small objects like pedestrians.

\par One possible solution towards the fore-mentioned issues is to replace the 2D object detector with 2D semantic segmentation and region-wise seed proposal with point-wise seed proposals. Intensive Point-based Object Detector (IPOD) \cite{yang2018ipod} by Yang et al. is a work in this direction. In the first step, 2D semantic segmentation is used to filter out back-ground points. This is achieved by projecting points to the image plane and associated point with 2D semantic labels. The resulting foreground point cloud retains the context information and fine-grained location, which is essential for the region-proposal and bbox regression. In the following point-wise proposal generation and bbox regression stage, two PointNet++ \cite{qi2017pointnet} based networks are used for proposal feature extraction and bbox prediction. In addition, a novel criterion called PointsIoU is proposed to speed up training and inference. This approach has yielded significant performance advantages over other state-of-the-art approaches in scenes with high occlusion or many objects.

\subsubsection{Multi-level Fusion}
\par Another possible direction of improvement is to combine result level fusion with feature level fusion, where one such work is PointFusion \cite{Xu_2018}. The PointFusion first utilizes an existing 2D object detector to generate 2D bboxes. These bboxes is used to select corresponding points, via projecting points to the image plane and locate points that pass through the bboxs. Finally, a ResNet \cite{He_2016} and a PointNet \cite{Charles_2017} based network combines image and point cloud features to estimate 3D objects. In this approach, image features and point cloud features are fused per-proposal for final object detection in 3D, which facilitates 3D bbox regression. However, its proposal stage is still amodal.  In SIFRNet \cite{zhao20193d}, frustum proposals are first generated from an image. Point cloud features in these frustum proposals are then combined with their corresponding image features for final 3D bbox regression. To achieve scale-invariant, the PointSIFT \cite{jiang2018pointsift} is incorporated into the network. Additionally, the SENet module is used to suppress less informative features.
% For cars the 3D bounding box IoU requirement is $70\%$. For pedestrians and cyclists the 3D bounding box IoU requirements are $50\%$.
\begin{table*}[]
\centering
\caption{Comparative results on KITTI 3D Object Detection benchmark (moderate difficulties)\cite{Geiger2012CVPR}.The IoU requirement for car, pedestrians and cyclists is $70\%$,  $50\%$ and  $50\%$ respectively.'FL','FT','RT' 'MS','PCR' stand for fusion level, fusion type, run-time, model size (number of parameters) and point cloud representation respectively.}
\label{tab:obj}
\centering
\begin{tabular}{|l|l|l|l|l|l|l|l|l|c|c|c|} 
\cline{1-12}
\multicolumn{2}{|l|}{\multirow{2}{*}{Methods}}                                                                                                                      & \multirow{2}{*}{FL}  & \multirow{2}{*}{FT} & \multirow{2}{*}{PCR} & \multirow{2}{*}{Models} & \multirow{2}{*}{Hardware} & \multirow{2}{*}{RT} & \multirow{2}{*}{MS} & Cars         & Pedestrians  & Cyclists      \\ 
\cline{10-12}
\multicolumn{2}{|l|}{}                                                                                                                                              &                                &                              &                      &             &              &        &                  & $AP_{3D}\%$  & $AP_{3D}\%$  & $AP_{3D}\%$   \\ 
\cline{1-12}
\multirow{13}{*}{\begin{tabular}[c]{@{}l@{}}Sequential \\ Models \end{tabular}} & \multirow{7}{*}{\begin{tabular}[c]{@{}l@{}}2D \\ Proposal \\ based \end{tabular}} & \multirow{5}{*}{Result }  & \multirow{5}{*}{ROI}         & Points               & F-PointNet \cite{Qi_2018}       & GTX1080    & 0.17s   & 7M         & 69.79        & 42.15        & 56.12         \\
                                                                                &                                                                                   &                                &                              & Points               & F-ConvNet \cite{Wang_2019} & GTX1080Ti  & 0.47s   & 30M        & 76.39        & 43.38        & 65.07         \\
                                                                                &                                                                                   &                                &                              & Voxels               & PC-CNN \cite{Du_2018}      & -          & -       & 276M       & 73.79        & -            & -             \\
                                                                                &                                                                                   &                                &                              & Points               & RoarNet \cite{Shin_2019}   & GTX1080Ti  & 0.1s    & -      & 73.04        & -            & -             \\
                                                                                &                                                                                   &                                &                              & Points               & IPOD \cite{yang2018ipod}   & GTX1080Ti  & 0.2s    & 24M           & 72.57        & 44.68        & 53.46         \\ 
\cline{3-12}
                                                                                &                                                                                   & Feature                   & Point                   & Multiple             & PointPainting \cite{vora2019pointpainting} & GTX1080 & 0.4s & -                   & 71.70        & 40.97        & 63.78         \\ 
\cline{3-12}
                                                                                &                                                                                   & \multirow{2}{*}{Multi}   & ROI                     & Points               & PointFusion \cite{Xu_2018}            & GTX1080  & 1.3s  & 6M                    & 63.00        & 28.04        & 29.42         \\
                                                                                &                                                                                   &                          & Point                   & Points               & SIFRNet \cite{zhao20193d}              & -  & -    & -                     & 72.05        & 60.85        & 60.34         \\ 
\cline{2-12}
                                                                                & \multirow{6}{*}{\begin{tabular}[c]{@{}l@{}}3D \\ Proposal \\ based \end{tabular}} & \multirow{5}{*}{Feature} & ROI                     & 2D views             & MV3D \cite{Chen_2017}                  & Titan X & 0.36s & 240M                  & 63.63        & -            & -             \\
                                                                                &                                                                                   &                                & ROI               & 2D views             & AVOD-FPN \cite{Ku_2018}                & Titan Xp  & 0.08s & 38M                    & 71.76        & 42.27        & 50.55         \\
                                                                                &                                                                                   &                                & ROI              & 2D views             & SCANet \cite{8682746}                   & GTX1080Ti  & 0.17s & -                    & 68.12        & 37.93        & 53.38         \\
                                                                                &                                                                                   &                                & Point                   & 2D views             & ContFuse \cite{Liang_2018_ECCV}  & GTX1080    & 0.06s & -                   & 68.78        & -            & -             \\
                                                                                &                                                                                   &                                & Point                   & Voxels             & MVX-Net \cite{Sindagi_2019}        & -       & -   & -                    & 77.43        & -            & -             \\ 
                                                                                &                                                                                   &                                & Pixel                   & 2D views             & BEVF \cite{8500387}              & -     & -     & -                  & -            & -            & 45.00         \\ 
\cline{3-12}
                                                                                &                                                                                   & Multi                    & Point                   & 2D views             & MMF \cite{Liang_2019_CVPR}             & GTX1080      & 0.08s & -                    & 77.43        & -            & -             \\ 
\cline{1-12}
 \multicolumn{2}{|l|}{ One-Step     Models }                                                                              &Feature  & Pixel                   & 2D views             & LaserNet++ \cite{meyer2019sensor}          & TITAN Xp    
 & 0.04s        & -             & -            & -            & -             \\
\cline{1-12}
\end{tabular}
\end{table*}

\subsubsection{Feature-level Fusion}
\par Early attempts \cite{Gupta_2014} \cite{Gupta_2016} of multimodal fusion are done in pixel-wise, where 3D geometry is converted to image format or appended as additional channels of an image. The intuition is to project 3D geometry onto the image plane and leverage mature image processing methods for feature extraction. The resulting output is also on the image plane, which is not ideal to locate objects in the 3D space. In 2014, Gupta et al. proposed DepthRCNN \cite{Gupta_2014},an R-CNN \cite{Girshick_2014} based architecture for 2D object detection, instance and semantic segmentation. It encodes 3D geometry from Microsoft Kinect camera in image's RGB channels, which are Horizontal disparity, Height above ground, and Angle with gravity (HHA). Gupta et al. extended Depth-RCNN \cite{7299105} in 2015 for 3D object detection by aligning 3D CAD models, yielding significant performance improvement. In 2016, Gupta et al. developed a novel technique for supervised knowledge transfer between networks trained on image data and unseen paired image modality (depth image) \cite{Gupta_2016}. In 2016, Schlosser et al. \cite{7487370} further exploited learning RGB-HHA representations on 2D CNNs for pedestrian detection. However, the HHA data are generated from the LiDAR's depth instead of a depth camera. The authors also noticed that better results can be achieved if the fusion of RGB and HHA happens at deeper layers of the network.

%  \par To locate objects accurately in 3D, current works often employs point-wise fusion. In this approach, image features are appended to each point in the point cloud.
% \par One of the challenges of point-wise fusion is the resolution mismatch between dense RGB and sparse depth. 

\par The resolution mismatch between dense RGB and sparse depth means only a small portion of pixels have corresponding points. Therefore, to directly append RGB information to points leads to the loss of most texture information, rendering the fusion pointless. To mitigate this challenge, PointPainting \cite{vora2019pointpainting} extract high-level image semantic before the per-point fusion. To be more specific, PointPainting \cite{vora2019pointpainting} follows the idea of projecting points to 2D semantic maps in \cite{yang2018ipod}. But instead of using 2D semantics to filter non-object points, 2D semantics is simply appended to point clouds as additional channels. The authors argued that this technique made PointPainting flexible as it enables any point cloud networks to be applied on this fused data. To demonstrate this flexibility, the fused point cloud is fed into multiple existing point cloud detectors, which are based on the PointRCNN \cite{Shi_2019}, the VoxelNet \cite{Zhou_2018} and the PointPillar \cite{Lang_2019}. However, this would lead to the coupling between image and LiDAR models. This requires the LiDAR model to be re-trained when the image model changes, which reduces the overall reliability and increases the development cost.

\begin{figure}
\centering
\includegraphics[scale=0.27]{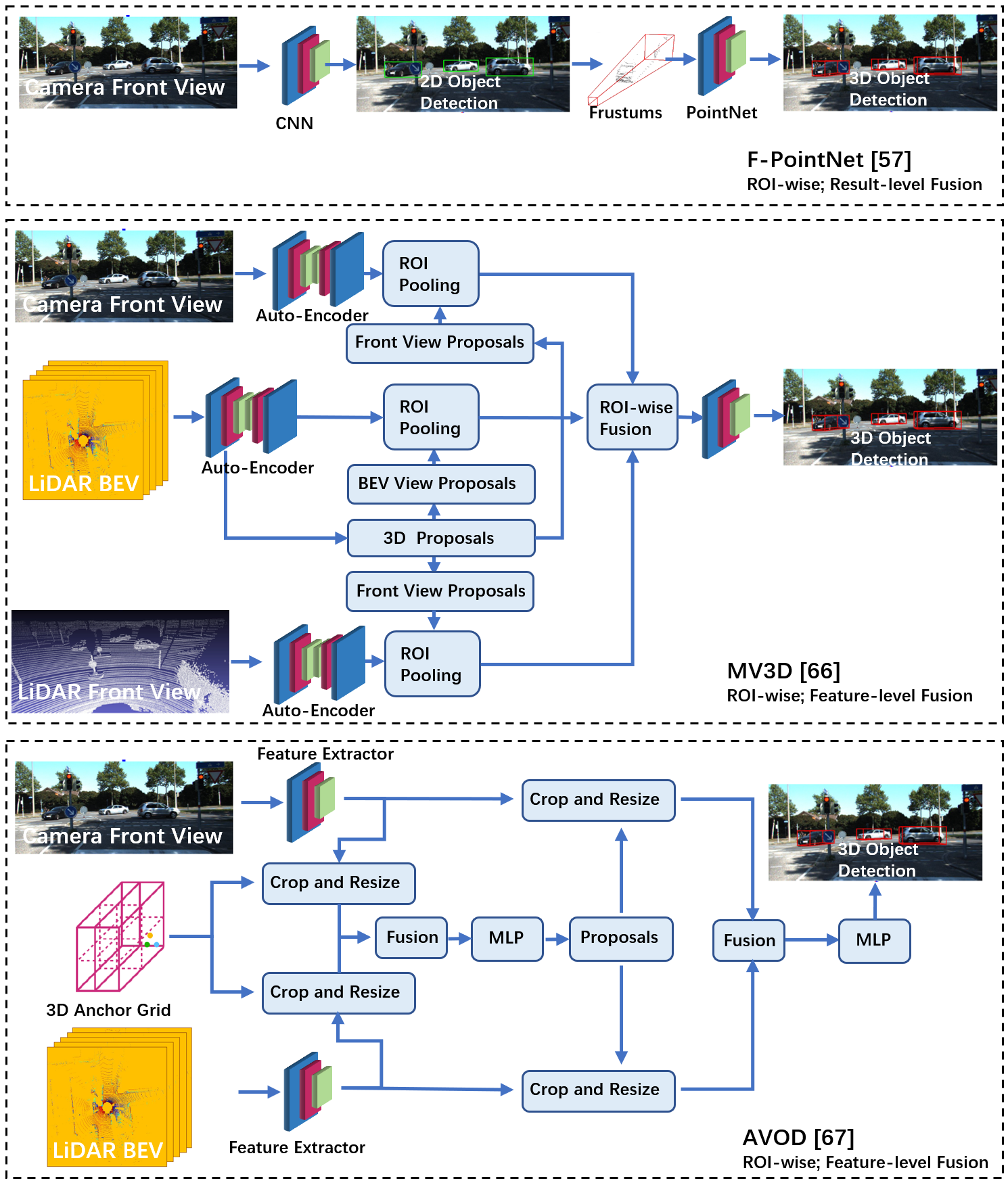}
\caption{A comparison between three typical model architectures for dynamic object detection.}
\label{arch2}
\end{figure}

\subsection{3D Proposal Based Sequential Model}

\par In a 3D proposal based sequential model, 3D proposals are directly generated from 2D or 3D data. The elimination of 2D to 3D proposal transformation greatly limits the 3D search space for 3D object detection. Common methods for 3D proposal generation includes the multi-view approach and the point cloud voxelization approach.

\par Multi-view based approach exploits the point cloud's bird's eye view (BEV) representation for 3D proposal generation. The BEV is the preferred viewpoint because it avoids occlusions and retains the raw information of objects' orientation and $x, y$ coordinates. These orientation and $x, y$ coordinates information is critical for 3D object detection while making coordinate transformation between BEV and other views straight-forward. 

% However, in this approach data fusion often happens at selected candidate regions in a high-level feature space, 

\par Point cloud voxelization transforms the continuous irregular data structure to a discrete regular data structure. This makes it possible to apply standard 3D discrete convolution and leverage existing network structures to process point cloud. The drawback is the loss of some spatial resolution, which might contain fine-grained 3D structure information.

\subsubsection{Feature-level fusion}

\par  One of the pioneering and most important works in generating 3D proposals from BEV representations is MV3D \cite{Chen_2017}. MV3D generate 3D proposals on pixelized top-down LiDAR feature map (height, density and intensity). These 3D candidates are then projected to the LiDAR front view and image plane to extract and fuse region-wise features for bbox regression. The fusion happens at the region-of-interest (ROI) level via ROI pooling. The $ROI_{views}$ of views is defined as:
\begin{equation}\label{eq7}
\mathrm{ROI}_{views}=\mathbf{T}_{3 \mathrm{D} \rightarrow views}\left(p_{3 \mathrm{D}}\right), views\in\{\mathrm{BV}, \mathrm{FV}, \mathrm{RGB}\}
\end{equation}
where $\mathbf{T}_{3 \mathrm{D} \rightarrow views}$ represents the transformation function that project point cloud $p_{3 \mathrm{D}}$ from 3D space to bird's eye view (BEV), front view (FV) and the image plane (RGB). The ROI-pooling $\mathbf{R}$ to obtain feature vector $f_{views}$ is defined as:
\begin{equation}\label{eq8}
f_{views}=\mathbf{R}\left(x, \mathrm{ROI}_{views}\right), views \in\{\mathrm{BV}, \mathrm{FV}, \mathrm{RGB}\}
\end{equation}
% \par Although MV3D out-performed the state-of-the-art models by a remarkable margin, there are a few flaws.

\par There are a few drawbacks of the MV3D. First, generating 3D proposals on BEV assumes that all objects of interest are captured without occlusions from this view-point. This assumption does not hold well for small object instances, such as pedestrians and bicyclists, which can be fully occluded by other large objects in the point cloud. Secondly, spatial information of small object instances is lost during the down-sample of feature maps caused by consecutive convolution operations. Thirdly, object-centric fusion combines feature maps of image and point clouds through ROI-pooling, which spoils fine-grained geometric information in the process. It is also worth noting that redundant proposals lead to repetitive computation in the bbox regression stage. To mitigate these challenges, multiple methods have been put forward to improve MV3D.

\begin{table*}[]
\centering
\caption{A summarization and comparison between dynamic object detection models}
\begin{tabular}{|m{18mm}|m{58mm}|m{30mm}|m{58mm}|}
\hline
 Methods: & Key features:                                                                                                                                                                                                                             & Advantages:                                                                                                                  & Disadvantages:                                                                                                                                                                                                                                                                       \\ \hline
 Frustum based \cite{Qi_2018} \cite{Wang_2019} \cite{Du_2018} \cite{Shin_2019} \cite{yang2018ipod} \cite{zhao20193d} \cite{Xu_2018} & \begin{tabular}[c]{p{55mm}p{55mm}}1.Leverage image object detector to generate 2D proposals and project them to form frustum 3D search spaces for 3D object detection.\\ 2.Result-level fusion / Multi-level fusion\end{tabular} & 3D search space is limited using 2D results to reduce computational cost.                                                   & \begin{tabular}[c]{p{55mm}p{55mm}}1. Due to sequential result-level fusion, the overall performance is limited by the image detector.\\ 2. Redundant information from multimodal sensors is not leveraged.\end{tabular}                                                                     \\ \hline
 Point-fusion based \cite{vora2019pointpainting}& \begin{tabular}[c]{p{55mm}p{55mm}}1.Fuse high-level image semantics point-wise and perform 3D object detection in the fused point cloud.\\ 2.Feature-level fusion.\end{tabular}                                                                   & Solve the resolution mismatch problem between dense RGB and sparse depth via fusing high-level image semantics to points.   & \begin{tabular}[c]{p{55mm}p{55mm}}1.Image and LiDAR models are highly coupled, which reduce overall reliability and increase development cost. \\ 2.3D search space is not limited which leads to high computational cost.\end{tabular}                                                      \\ \hline
Multi-view based \cite{Chen_2017} \cite{Ku_2018} \cite{8682746} \cite{Liang_2018_ECCV} \cite{Liang_2019_CVPR} & \begin{tabular}[c]{p{55mm}p{55mm}}1.Generate 3D proposals from BEV and perform 3D bboxes regression on these proposals.\\ 2.Feature-level fusion.\end{tabular}                                                                                    & Enable the use of standard 2D convolution and off-the-shelf models, making it much more scalable. & \begin{tabular}[c]{p{55mm}p{55mm}}1.Assume all objects are visible in LiDAR's BEV, which is often not the case. \\ 2.Spatial information of small object instances is lost during consecutive convolution operations\\ 3.ROI fusion spoils fine-grained geometric information.\end{tabular} \\ \hline
Voxel-based \cite{Sindagi_2019} \cite{10.1007/978-3-319-10599-4_41} & \begin{tabular}[c]{p{55mm}p{55mm}}1.Fuse image semantics voxel-wise and perform 3D bboxes regression using voxel-based network.\\ 2.Feature-level fusion.\end{tabular}                                                                                     & Compatible with standard 3D convolution.                                                                                     & \begin{tabular}[c]{p{55mm}p{55mm}}1. Spatial resolution and fine-grained 3D geometry information are lost during voxelization.\\ 2. Computation and memory footprint to grow cubically with the resolution, making it unscalable.\end{tabular}                                                \\ \hline
\end{tabular}
\end{table*}

\par To improve the detection of small objects, the Aggregate View Object Detection network (AVOD) \cite{Ku_2018} first improved the proposal stage in MV3D \cite{Chen_2017} with feature maps from both BEV point cloud and image. Furthermore, an auto-encoder architecture is employed to up-sample the final feature maps to its original size. This alleviates the problem that small objects might get down-sampled to one 'pixel' with consecutive convolution operations. The proposed feature fusion Region Proposal Network (RPN) first extracts equal-length feature vectors from multiple modalities (BEV point cloud and image) with crop and resize operations. Followed by a $1\times1$ convolution operation for feature space dimensionality reduction, which can reduce computational cost and boost up speed. Lu et al.\cite{8682746} also utilized an encoder-decoder based proposal network with Spatial-Channel Attention (SCA) module and Extension Spatial Upsample (ESU) module. The SCA can capture multi-scale contextual information, whereas ESU recovers the spatial information.

% \textbf{Point-wise fusion}
\par  One of the problems in object-centric fusion methods \cite{Ku_2018}\cite{Chen_2017} is the loss of fine-grained geometric information during ROI-pooling. The ContFuse \cite{Liang_2018_ECCV} by Liang et al. tackles this information lost with point-wise fusion. This point-wise fusion is achieved with continuous convolutions \cite{wang2018deep} fusion layers that bridge image and point cloud features of different scales at multiple stages in the network. This is achieved by first extracting K-nearest neighbour points for each pixel in the BEV representation of point cloud. These points are then projected to the image plane to retrieve related image features. Finally, the fused feature vector is weighted according to their geometry offset to the target 'pixel' before feeding into MLPs. However, point-wise fusion might fail to take full advantage of high-resolution images when the LiDAR points are sparse. In \cite{Liang_2019_CVPR} Liang et al. further extended point-wise fusion by combining multiple fusion methodologies, such as signal-level fusion (RGB-D), feature-level fusion, multi-view and depth completion. In particular, depth completion upsamples sparse depth map using image information to generate a dense pseudo point cloud. This up-sampling process alleviates the sparse point-wise fusion problem, which facilitates the learning of cross-modality representations. Furthermore, the authors argued that multiple complementary tasks (ground estimation, depth completion and 2D/3D object detection) could assists the network achieve better overall performance. However, point-wise/pixel-wise fusion leads to the 'feature blurring' problem. This 'feature blurring' happens when one point in the point cloud is associated with multiple pixels in the image or the other way around, which confound the data fusion. Similarly, wang et al. \cite{8500387} replace the ROI-pooling in MV3D \cite{Chen_2017} with sparse non-homogeneous pooling, which enables effective fusion between feature maps from multiple modalities.

\par  MVX-Net \cite{Sindagi_2019} presented by Sindagi et al. introduced two methods that fuse image and point cloud data point-wise or voxel-wise. Both methods employ a pre-trained 2D CNN for image feature extraction and a VoxelNet \cite{Zhou_2018} based network to estimate objects from the fused point cloud. In the point-wise fusion method, the point cloud is first projected to image feature space to extract image features before voxelization and processed by VoxelNet. The voxel-wise fusion method first voxelized the point cloud before projecting non-empty voxels to the image feature space for voxel/region-wise feature extraction. These voxel-wise features are only appended to their corresponding voxels at a later stage of the VoxelNet. MVX-Net achieved state-of-the-art results and out-performed other LiDAR-based methods on the KITTI benchmark while lowering false positives and false negatives rate compared to \cite{Zhou_2018}.

% \textbf{Voxel-wise fusion}
\par The simplest means to combine the voxelized point cloud and image is to append RGB information as additional channels of a voxel. In a 2014 paper by Song et al. \cite{10.1007/978-3-319-10599-4_41} 3D object detection is achieved by sliding a 3D detection window on the voxelized point cloud. The classification is performed by an ensemble of Exemplar-SVMs. In this work, color information is appended to voxels by projection. Song et al. further extended this idea with 3D discrete convolutional neural networks \cite{song2016deep}. In the first stage, the voxelized point cloud (generated from RGB-D data) is first processed by Multi-scale 3D RPN for 3D proposal generation. These candidates are then classified by joint Object Recognition Network (ORN), which takes both image and voxelized point cloud as inputs. However, the volumetric representation introduces boundary artifacts and spoils fine-grained local geometry. Secondly, the resolution mismatch between image and voxelized point cloud makes fusion inefficient.

\subsection{One-step Models}

\par One-step models perform proposal generation and bbox regression in a single stage. By fusing the proposal and bbox regression stage into one-step, these models are often more computationally efficient. This makes them more well-suited for real-time applications on mobile computational platforms.

\par Meyer et al. \cite{meyer2019sensor} extended the LaserNet \cite{Meyer_2019} to multi-task and multimodal network, performing 3D object detection and 3D semantic segmentation on fused image and LiDAR data. Two CNN process depth-image (generated from point cloud) and front-view image in a parallel manner and fuse them via projecting points to the image plane to associate corresponding image features. This feature map is fed into the LaserNet to predict per-point distributions of the bounding box and combine them for final 3D proposals. This method is highly efficient while achieving state-of-the-art performance.

\section{Stationary Road Object Detection}

% \par Stationary road object detection is one of the fundamental challenges and crucial prerequisites for autonomous driving.
\par This section focuses on reviewing recent advances in camera-LiDAR fusion based stationary road object detection methods. Stationary road objects can be categorized into on-road objects (e.g. road surfaces and road markings) and off-road objects (e.g. traffic signs). On-road and off-road objects provide regulations, warning bans and guidance for autonomous vehicles. 

% Recent studies have proposed to utilize both LiDAR and camera for accurate drivable area (road/lane) detection
\par In Figure 6 \& 7,  typical model architecture in lane/road detection and traffic sign recognition (TSR) are compared. Table IV presents the comparative results of different models on the KITTI road benchmark \cite{Geiger2012CVPR} and gives summarization and comparison between these models.

\begin{table*}[]
\centering
\caption{Comparative results on KITTI road benchmark \cite{Geiger2012CVPR} with summarization and comparison between lane/road detection models.'MaxF (\%)' stands for Maximum F1-measure on KITTI Urban Marked Road test set. 'FL','MS' stand for Fusion Level and model parameter size.}
\begin{tabular}{|m{17mm}|m{20mm}|m{9mm}|m{8mm}|m{7mm}|m{50mm}|m{40mm}|}
\hline
Method:      & Inputs:                   & FL:   &MaxF  &MS & Highlights:                                                               & Disadvantages:                                        \\ \hline
\multicolumn{7}{|c|}{BEV-based Lane/Road Detection}                                                                                                                                    \\ \hline
DMLD \cite{8594388}        & BEV Image + BEV Point Cloud & Feature & -      & 50M   & Predict dense BEV height map from sparse depth which improves performance & Cannot distinguish between different lane types       \\ \hline
EDUNet \cite{8500549}       & Fused BEV Occupation Grid & Signal   & 93.81  & 143M  & Signal-level fusion reduces model complexity                              & Loss of dense RGB/texture information in early fusion \\ \hline
TSF-FCN \cite{8500551}     & BEV Image + BEV Point Cloud & Feature & 94.86  & 2M      & Relatively moderate computational cost                                    & Relatively moderate performance                       \\ \hline
MSRF \cite{8813983}         & BEV Image + BEV Point Cloud & Feature & 96.15  & - & Multi-stage fusion at different network depth improves performance        & Relatively high computational cost                    \\ \hline
\multicolumn{7}{|c|}{Camera-view based Lane/Road Detection}                                                                                                                            \\ \hline
LidCamNet \cite{Caltagirone_2019}       & Image + Dense Depth Image & Multi & 96.03 & 3M & Explore and compared multiple fusion strategies for road detection        & Rely on image for depth up-sampling       \\ \hline
PLARD \cite{8707128}       & Image + Dense Depth Image   & Feature  & 97.05 & - & Progressive fusion strategy improves performance                          & Relatively high computational cost       \\ \hline
SpheSeg \cite{9084150} & Image + Depth Image        & Feature & 96.94 & 1.4M & Spherical coordinate transformation reduces input size and improves speed & Road detection in 2D space rather than 3D space       \\ \hline
\end{tabular}
\end{table*}

\subsection{Lane/Road detection}
\par Existing surveys \cite{ma2018mobile} \cite{narote2018review} \cite{xing2018advances} have presented detailed reviews on traditional multimodal road detection methods. These methods \cite{huang2009finding} \cite{6856454} \cite{xiao2018hybrid} \cite{7225685} mainly rely on vision for road/lane detection while utilizing LiDAR for the curb fitting and obstacle masking. Therefore, this section focuses on recent advances in deep learning based fusion strategies for road extraction.

\begin{figure}
\centering
\includegraphics[scale=0.3]{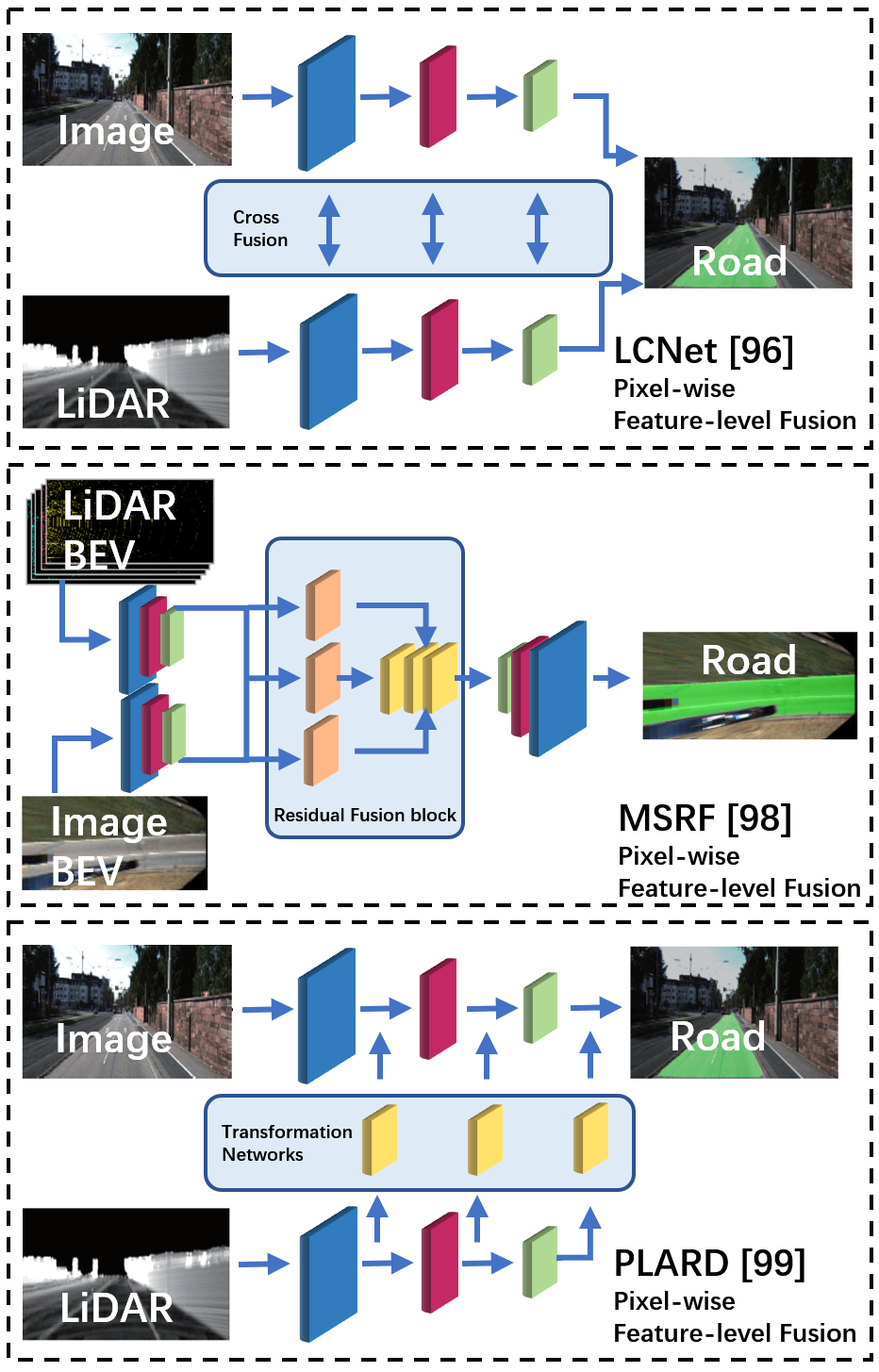}
\caption{Some typical model architectures and fusion methods for road/lane detection}
\label{archsemantic}
\end{figure}

\par Deep leaning based road detection methods can be grouped into BEV-based or front-camera-view-based. BEV-based methods \cite{8594388} \cite{8500551} \cite{8813983} \cite{8500549} project LiDAR depth and images to BEV for road detection, which retains the objects' original $x, y$ coordinates and orientation. In \cite{8594388}, the dense BEV height estimation is predicted from the point cloud using a CNN, which is then fused with the BEV image for accurate lane detection. However, this method cannot distinguish different lane types. Similarly, Lv et al. \cite{8500551} also utilized the BEV LiDAR grid map and the BEV image but instead processed them in a parallel manner. Yu et al. \cite{8813983} proposed a multi-stage fusion strategy (MSRF) that combines image-depth features at different network levels, which significantly improves its performance. However, this strategy also relatively increases its computational cost. Wulff et al. \cite{8500549} used signal-level fusion to generate a fused BEV occupation grid, which is processed by a U-net based road segmentation network. However, the signal-level fusion between dense RGB and sparse depth leads to the loss of dense texture information due to the low grid resolution. 
\par Front-camera-view-based methods \cite{Caltagirone_2019} \cite{8707128} \cite{9084150}  project LiDAR depth to the image plane to extract road surface, which suffers from accuracy loss in the translation of 2D to 3D boundaries. The LCNet \cite{Caltagirone_2019} compared signal-level fusion (early fusion) and feature level fusion (late and cross fusion) for road detection, which finds the cross fusion is the best performing fusion strategy. Similar to \cite{8500551}, PLARD \cite{8707128} fuses image and point cloud features progressively in multiple stages. Lee et al. \cite{9084150} focused on improving speed via a spherical coordinate transformation scheme that reduces the input size. This transformed camera and LiDAR data are further processed by a SegNet based semantic segmentation network. 

\subsection{Traffic sign recognition}

\begin{figure}
\centering
\includegraphics[scale=0.28]{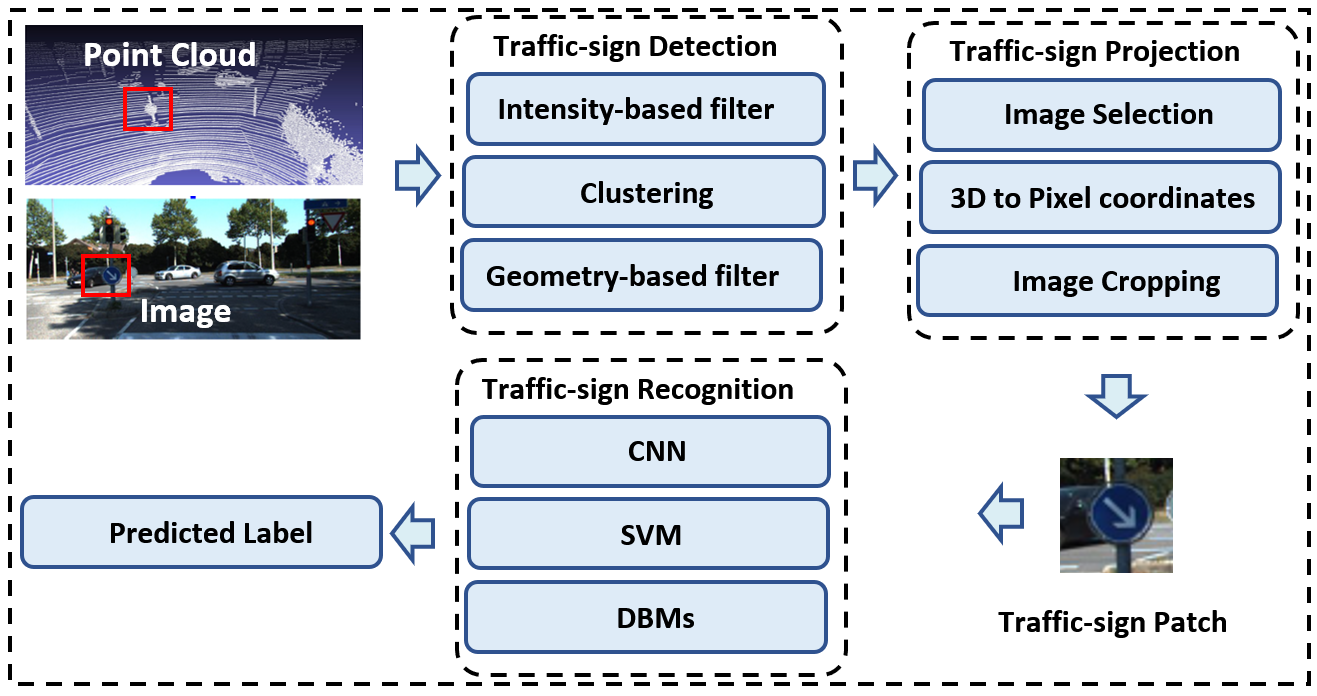}
\caption{A typical fusion-based traffic-sign recognition pipeline.}
\label{trafficsign}
\end{figure}

\par In LiDAR scans, traffic-signs are highly distinguishable due to its retro-reflective property, but the lack of dense texture makes it difficult to classify. On the contrary, traffic sign image patches can be easily classified. However, it is difficult for vision-based TSR system to locate these traffic-signs in the 3D space. Therefore, various studies have proposed to utilize both camera and LiDAR for TSR. Existing reviews \cite{ma2018mobile} \cite{mogelmose2012vision} have comprehensively covered traditional traffic sign recognition methods and part of the deep learning methods. Hence, this section presents a brief overview of traditional traffic sign recognition methods and mostly focuses on recent advances. In a typical TSR fusion pipeline \cite{7378972} \cite{yu2016bag} \cite{SOILAN201692} \cite{arcos2017exploiting} \cite{8848423}, traffic-signs are first located in the LiDAR scan based on its retro-reflective property. These 3D positions of detected traffic-signs are then projected to the image plane to generate traffic-signs patches, which are fed into an image classifier for classification. This TSR fusion pipeline is shown in Fig. 7.

\par For methods that employ the typical TSR fusion pipeline, the main difference is on the classifier. These classifiers include deep Boltzmann machine (DBMs) based hierarchical classifier\cite{yu2016bag}, SVMs \cite{7378972} and DNN \cite{arcos2017exploiting}. To summarize, these methods all employ result-level fusion and a hierarchical object detection model. They assume traffic-signs are visible in the LiDAR scan, which sometimes is not the case due to the occlusion. Furthermore, this pipeline is limited by the detection range of mobile LiDARs.

\par To mitigate these challenges, Deng et al. \cite{8013749} combined image and point cloud to generate a colorized point cloud for both traffic-sign detection and classification. In addition, the 3D geometrical properties of detected traffic-signs are utilized to reduce false-positives. In \cite{8325411} traffic-signs are detected based on prior knowledge, which includes road geometry information and traffic-sign geometry information. The detected traffic-signs patches are classified by a Gaussian–Bernoulli DBMs model. Following this ideal, Guan et al. \cite{8848423} further improved the traffic sign recognition part using a convolutional capsule network. To summarize, these methods improve the traffic sign detection stage with multimodal data and prior knowledge. However, prior knowledge is often region-specific, which makes it difficult to generalize to other parts of the world.

%  In a 2016 study \cite{SOILAN201692} by Soilán st al., the point cloud is first segmented via ground point filtering and intensity filtering. These segments are further processed to locate individual traffic-sign and their locations are projected to the image plane. The detected traffic-sign patches are described using the histogram of oriented gradients (HOG) and classified via hierarchical SVM models.

\begin{figure}
\centering
\includegraphics[scale=0.23]{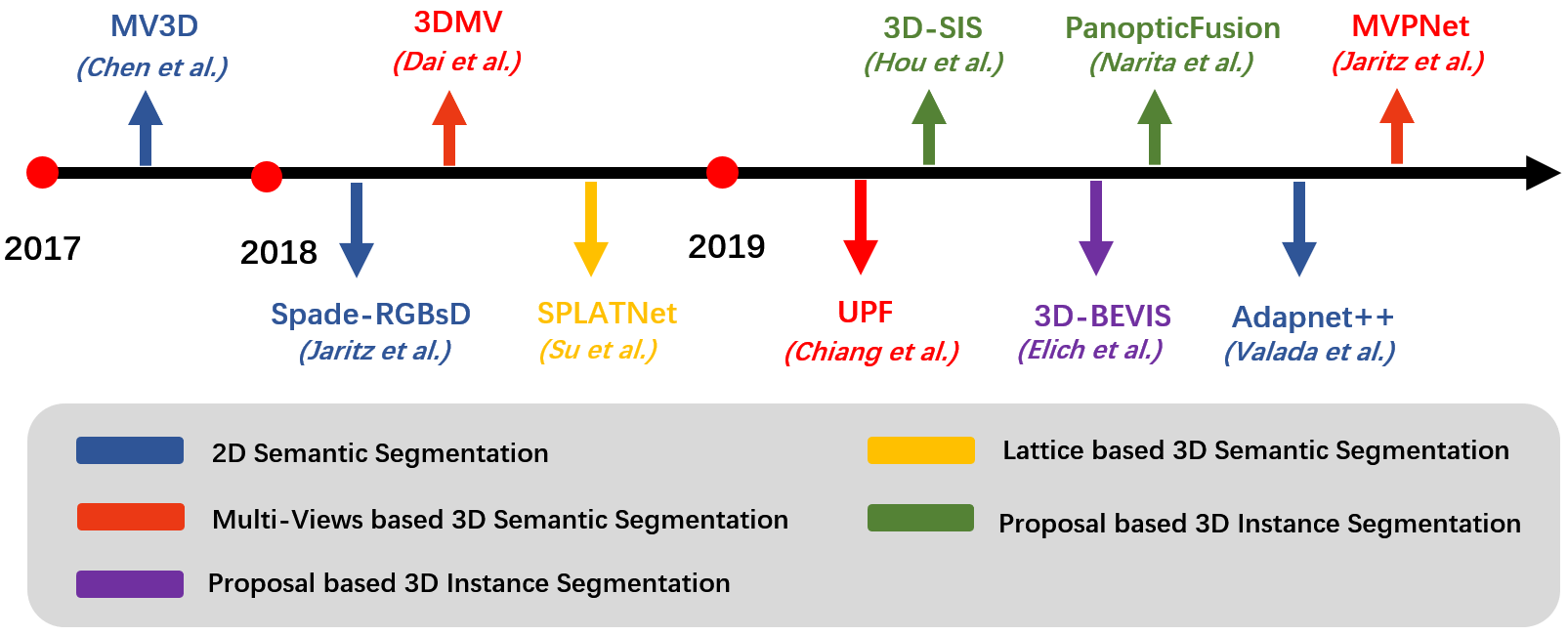}
\caption{ Timeline of 3D semantic segmentation networks and their corresponding fusion levels. }
\label{timeline2}
\end{figure}

\section{Semantic Segmentation}
\par This section reviews existing camera-LiDAR fusion methods for 2D semantic segmentation, 3D semantic segmentation and instance segmentation. 2D/3D semantic segmentation aims to predict per-pixel and per-point class labels, while instance segmentation also cares about individual instances. Figure 8 \& 9 present a timeline of 3D semantic segmentation networks and typical model architectures.

\subsection{2D Semantic Segmentation}
\subsubsection{Feature-level fusion}
\par Sparse \& Dense \cite{Jaritz_2018} presented a NASNet \cite{Zoph_2018} based auto-encoder network that can be used for 2D semantic segmentation or depth completion leveraging image and sparse depths. The image and corresponding sparse depth map are processed by two parallel encoders before fused into the shared decoder. Valada et al. \cite{Valada_2019} employed a multi-stage feature-level fusion of varying depth to facilitate semantic segmentation. Caltagirone et al.\cite{Caltagirone_2019} utilized up-sampled depth-image and image for 2D semantic segmentation. This dense depth-image is up-sampled using sparse depth-images (from point cloud) and images \cite{inproceedings}. The best performing cross-fusion model processes dense depth-image and image data in two parallel CNN branches with skip-connections in between and fuses the two feature maps in the final convolution layer.

\subsection{3D Semantic Segmentation}

\subsubsection{Feature-level fusion}
\par Dai et al.\cite{Dai_2018} presented 3DMV, a multi-view network for 3D semantic segmentation which fuse image semantic and point features in voxelized point cloud. Image features are extracted by 2D CNNs from multiple aligned images and projected back to the 3D space. These multi-view image features are max-pooled voxel-wise and fused with 3D geometry before feeding into the 3D CNNs for per-voxel semantic prediction. 3DMV out-performed other voxel-based approaches on ScanNet \cite{Dai_2017} benchmark. However, the performance of voxel-based approaches is determined by the voxel-resolution and hindered by voxel boundary artifacts. 

\par To alleviate problems caused by point cloud voxelization, Chiang et al. \cite{Chiang_2019} proposed a point-based semantic segmentation framework (UPF) that also enables efficient representation learning of image features, geometrical structures and global context priors. Features of rendered multi-view images are extracted using a semantic segmentation network and projected to 3D space for point-wise feature fusion. This fused point cloud is processed by two PointNet++ \cite{qi2017pointnet} based encoders to extract local and global features before feeding into a decoder for per-point semantic label prediction. Similarly, Multi-View PointNet (MVPNet) \cite{jaritz2019multiview} fused multi-view images semantics and 3D geometry to predict per-point semantic labels.

% \subsubsection{Lattice based} 

\par Permutohedral lattice representation is an alternative approach for multimodal data fusion and processing. Sparse Lattice Networks (SPLATNet) by Su et al. \cite{Su_2018} employed sparse bilateral convolution to achieve spatial-aware representation learning and multimodal (image and point cloud) reasoning. In this approach, point cloud features are interpolated onto a $d_l$-dimensional permutohedral lattice, where bilateral convolution is applied. The results are interpolated back onto the point cloud. Image features are extracted from multi-view images using a CNN and projected to the 3D lattice space to be combined with 3D features. This fused feature map is further processed by CNN to predict the per-point label.

\begin{figure}
\centering
\includegraphics[scale=0.26]{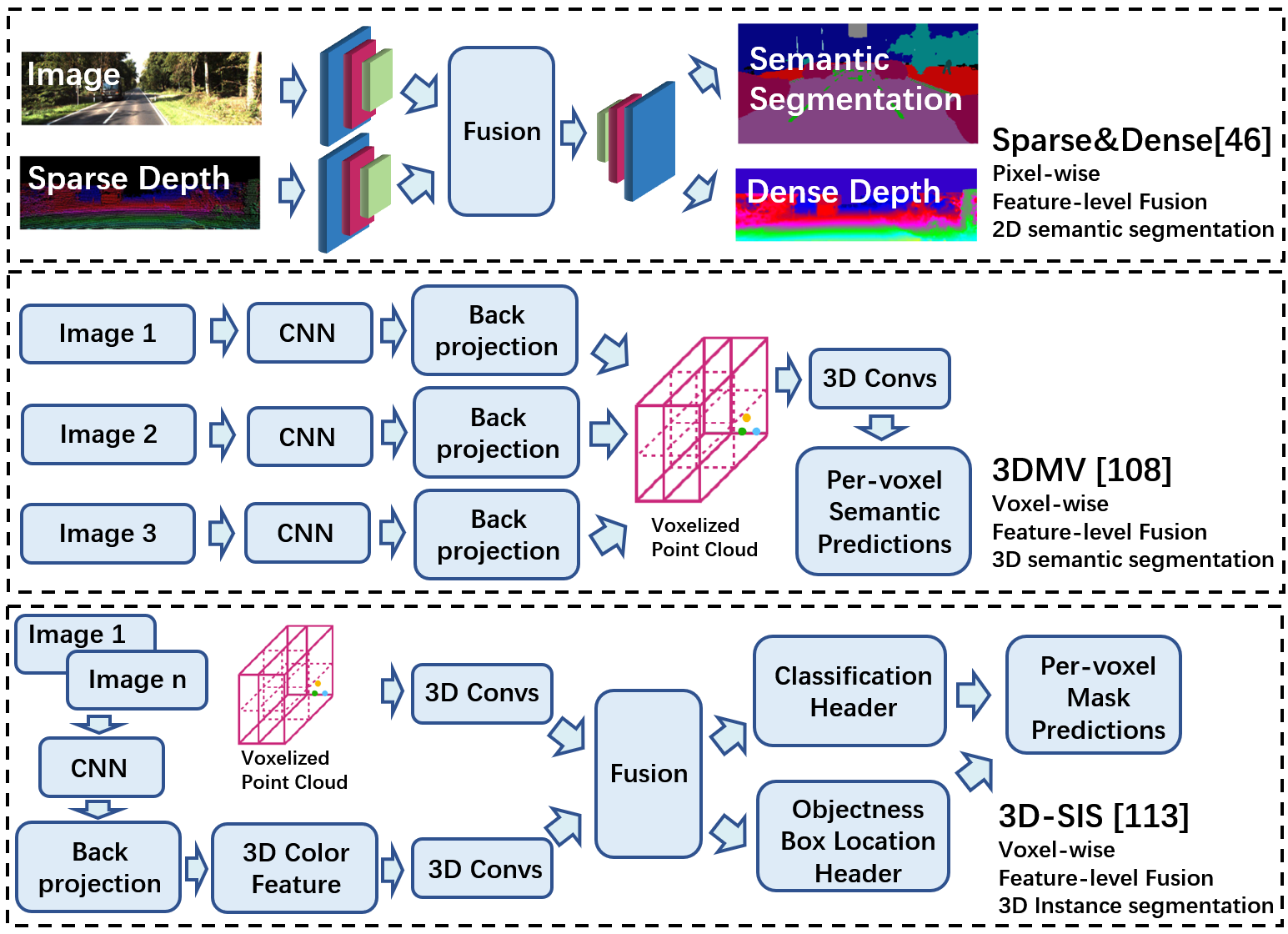}
\caption{Some typical model architectures and fusion methods for semantic segmentation}
\label{archsemantic}
\end{figure}

\subsection{Instance Segmentation}
\par In essence, instance segmentation aims to perform semantic segmentation and object detection jointly. It extends the semantic segmentation task by discriminating against individual instances within a class, which makes it more challenging.

\subsubsection{Proposal based} 
 Hou et al. presented 3D-SIS \cite{Hou_2019}, a two-stage 3D CNN that performs voxel-wise 3D instance segmentation on multi-view images and RGB-D scan data. In the 3D detection stage, multi-view image features are extracted and down-sampled using ENet \cite{paszke2016enet} based network. This down-sample process tackles the mismatch problem between a high-resolution image feature map and a low-resolution voxelized point cloud feature map. These down-sampled image feature maps are projected back to 3D voxel space and append to the corresponding 3D geometry features, which are then fed into a 3D CNN to predict object classes and 3D bbox poses. In the 3D mask stage, a 3D CNN takes images, point cloud features and 3D object detection results to predict per-voxel instance labels.

% \subsubsection{Proposal based} 
Narita et al. \cite{Narita_2019} extended 2D panoptic segmentation to perform scene reconstruction, 3D semantic segmentation and 3D instance segmentation jointly on RGB images and depth images. This approach takes RGB and depth frames as inputs for instance and 2D semantic segmentation networks. To track labels between frames, these frame-wise predicted panoptic annotations and corresponding depth are referenced by associating and integrating to the volumetric map. In the final step, a fully connected conditional random field (CRF) is employed to fine-tune the outputs. However, this approach does not support dynamic scenes and are vulnerable to long-term post drift.

\subsubsection{Proposal-free based}
 Elich et al. \cite{Elich_2019} presented 3D-BEVIS, a framework that performs 3D semantic and instance segmentation tasks jointly using the clustering method on points aggregated with 2D semantics. 3D-BEVIS first extract global semantic scores map and instance feature map from 2D BEV representation (RGB and height-above-ground). These two semantic maps are propagated to points using a graph neural network. Finally, the mean shift algorithm \cite{comaniciu2002mean} use these semantic features to cluster points into instances. This approach is mainly constraint by its dependence on semantic features from BEV, which could introduce occlusions from sensor displacements.

\begin{table*}[]
\centering
\caption{Comparative results on KITTI multi-object tracking benchmark (car) \cite{Geiger2012CVPR}. MOTA stands for multiple object tracking accuracy. MOPT stands for multiple object tracking precision. MT stands for Mostly tracked. ML stands for Mostly Lost. IDS stands for number of ID switches. FRAG stands for number of Fragments.}
\label{tab:track}
\centering
\begin{tabular}{|l|l|l|l|l|ccccll|} 
\cline{1-11}
Methods                 & Data-association & Models                             &Hardware  & Run-time                  & MOTA(\%)                   & MOTP(\%)                   & MT(\%)                       & ML(\%)                      & IDS   & FRAG                      \\ 
\cline{1-11}
\multirow{3}{*}{DBT~ ~} & min-cost ﬂow~    & DSM \cite{Frossard_2018}           &  GTX1080Ti                       & 0.1s                      & 76.15                     & 83.42                     & 60.00                       & 8.31                        & 296   & 868                       \\
                        & min-cost ﬂow~    & mmMOT \cite{Zhang_2019}            &  GTX1080Ti                       & 0.02s                     & 84.77                     & 85.21                     & 73.23                       & 2.77                        & 284   & 753                       \\
                        & Hungarian~       & MOTSFusion \cite{Luiten_2020}      &  GTX1080Ti                       & 0.44s                     & 84.83                     & 85.21                     & 73.08                       & 2.77                        & 275   & 759                       \\ 
\cline{1-11}
DFT                     & RFS              & Complexer-YOLO \cite{simon2019complexeryolo}    &  GTX1080Ti                 &0.01s                       &75.70                     & 78.46                     &58.00                         &5.08                       & 1186  &2092                 \\
\cline{1-11}
\end{tabular}

\end{table*}

\section{Objects Tracking}
\par Multiple object tracking (MOT) aims to maintain objects identities and track their location across data frames (over time), which is indispensable for the decision making of autonomous vehicles. To this end, this section reviews camera-LiDAR fusion based object tracking methods. Based on object initialization methods, MOT algorithms can be catalogued into Detection-Based Tracking (DBT) and Detection-Free Tracking (DFT) frameworks. DBT or Tracking-by-Detection framework leverages a sequence of object hypotheses and higher-level cues produced by an object detector to track objects. In DBT, objects are tracked via data (detections sequence) association or Multiple Hypothesis Tracking. On the contrary, the DFT framework is based on finite set statistics (FISST) for state estimation. Common methods include Multi-target Multi-Bernoulli (MeMBer) filter and Probability Hypothesis Density (PHD) filter.
\par Table V presents the performance of different models on the KITTI multi-object tracking benchmark (car) \cite{Geiger2012CVPR}. Figure 10 provides a comparison between DBT and DFT approaches.

\begin{figure}
\centering
\includegraphics[scale=0.3]{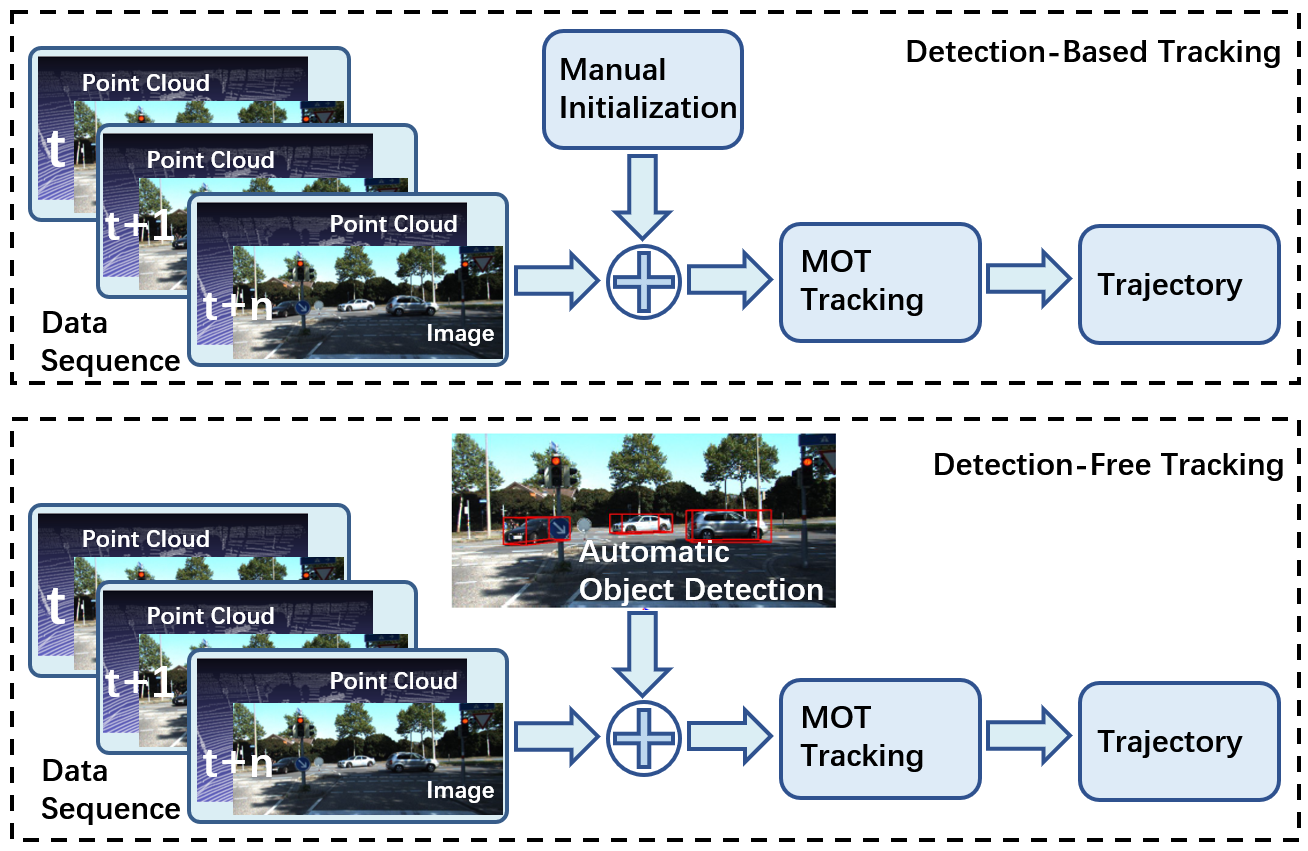}
\caption{A comparison between Detection-Based Tracking (DBT) and Detection-Free Tracking (DFT) approaches.}
\label{archT}
\end{figure}

\subsection{Detection-Based Tracking (DBT)}

\par The tracking-by-Detection framework consists of two stages. In the first stage, objects of interest are detected. The second stage associates these objects over time and formulates them into trajectories, which are formulated as linear programs. Frossard et al. \cite{Frossard_2018} presented an end-to-end trainable tracking-by-detection framework comprise of multiple independent networks that leverage both image and point cloud. This framework performs object detection, proposal matching and scoring, linear optimization consecutively. To achieve end-to-end learning, detection and matching are formulated via a deep structured model (DSM). Zhang et al. \cite{Zhang_2019} presented a sensor-agnostic framework, which employs a loss-coupling scheme for image and point cloud fusion. Similar to \cite{Frossard_2018}, the framework consists of three stages, object detection, adjacency estimation and linear optimization. In the object detection stage, image and point cloud features are extracted via a VGG-16 \cite{Simonyan15} and a PointNet \cite{qi2017pointnet} in parallel and fused via a robust fusion module. The robust fusion module is designed to work with both a-modal and multi-modal inputs. The adjacency estimation stage extends min-cost flow to multi-modality via adjacent matrix learning. Finally, an optimal path is computed from the min-cost flow graph. 
% \subsection{Tracking-by-Reconstruct}

\par Tracking and 3D reconstruct tasks can be performed jointly. Extending this idea, Luiten et al. \cite{Luiten_2020} leveraged 3D reconstruction to improve tracking, making tracking robust against complete occlusion. The proposed MOTSFusion consists of two stages. In the first stage, detected objects are associate with spatial-temporal tracklets. These tracklets are matched and merged into trajectories using the Hungarian algorithm. Furthermore, MOTSFusion can work with LiDAR mono and stereo depth.

\subsection{Detection-Free Tracking (DFT)}
\par In DFT objects are manually initialized and tracked via filtering based methods. The complexer-YOLO \cite{simon2019complexeryolo} is a real-time framework for decoupled 3D object detection and tracking on image and point cloud data. In the 3D object detection phase, 2D semantics are extracted and fused point-wise to the point cloud. This semantic point cloud is voxelized and fed into a 3D complex-YOLO for 3D object detection. To speed up the training process, IoU is replaced by a novel metric called Scale-Rotation-Translation score (SRTs), which evaluates 3 DoFs of the bounding box position. Multi-object tracking is decoupled from the detection, and the inference is achieved via Labeled Multi-Bernoulli Random Finite Sets Filter (LMB RFS).

\section{Online Cross-Sensor Calibration}

\par One of the preconditions of the camera-LiDAR fusion pipeline is a flawless registration/calibration between sensors, which can be difficult to satisfy. The calibration parameters between sensors change constantly due to mechanical vibration and heat fluctuation. As most fusion methods are extremely sensitive towards calibration errors, this could significantly cripple their performance and reliability. Furthermore, offline calibration is a troublesome and time-consuming procedure. Therefore, studies on online automatic cross-sensor calibration have significant practical benefits.

% \par Online calibration methods have to perform in natural settings without a calibration target. Many studies \cite{6631094} \cite{taylor2013automatic} \cite{pandey2015automatic} \cite{miled2016hybrid} leverage mutual information (MI) between different modalities to estimate extrinsics. MI-based approaches attempt to find a set of extrinsics that maximizes mutual information (raw intensity value or edge intensity) between image and depth image. However, MI-based methods are not robust against texture-rich environments, large de-calibrations and occlusions caused by sensor displacements. 
% To this end, LiDAR-enabled visual-odometry based method \cite{chien2016visual} use the camera's ego-motion to estimate and evaluate camera-LiDAR extrinsic parameters. Nevertheless, \cite{chien2016visual} still struggles with large de-calibrations and cannot run in real-time.
% \cite{6631094,taylor2013automatic,pandey2015automatic}
\subsection{Classical Online Calibration}
\par Online calibration methods estimate extrinsic in natural settings without a calibration target. Many studies \cite{6631094} \cite{taylor2013automatic} \cite{pandey2015automatic} \cite{miled2016hybrid} find extrinsic by maximizing mutual information (MI) (raw intensity value or edge intensity) between different modalities. However, MI-based methods are not robust against texture-rich environments, large de-calibrations and occlusions caused by sensor displacements. Alternatively, the LiDAR-enabled visual-odometry based method \cite{chien2016visual} uses the camera's ego-motion to estimate and evaluate camera-LiDAR extrinsic parameters. Nevertheless, \cite{chien2016visual} still struggles with large de-calibrations and cannot run in real-time.

\subsection{DL-based Online Calibration}
\par To mitigate the aforementioned challenges, Schneider et al. \cite{Schneider_2017} designed a real-time capable CNN (RegNet) to estimate extrinsic, which is trained on randomly decalibrated data. The proposed RegNet extracts image and depth feature in two parallel branches and concatenates them to produce a fused feature map. This fused feature map is fed into a stack of Network in Network (NiN) modules and two fully connected layers for feature matching and global regression. However, the RegNet is agnostic towards the sensor's intrinsic parameters and needs to be retrained once these intrinsic changes. To solve this problem, the CalibNet \cite{Iyer2018CalibNetGS} learns to minimize geometric and photometric inconsistencies between the miscalibrated and target depth in a self-supervised manner. Because intrinsics are only used during the 3D spatial transformer, the CalibNet can be applied to any intrinsically calibrated cameras. Nevertheless, deep learning based cross-sensor calibration methods are computationally expensive.

\begin{table*}[]
\centering
\caption{Open challenges related to performance improvement, reliability enhancement.}
\begin{tabular}{|m{70mm}|m{90mm}|}
\hline
Open Challenges:                                      & Possible solutions / directions:                  \\ \hline
\multicolumn{2}{|c|}{Performance-related Open Research Questions}                                         \\ \hline
What should be the data representation of fused data? & Point Representation + Point Convolution          \\ \hline
How to encode Temporal Context?                       & RNN/LSTM + Generative Models                      \\ \hline
What should be the learning scheme?                   & Unsupervised + Weakly-supervised learning         \\ \hline
When to use deep learning methods?                    & On applications with explicit objectives that can be verified objectively         \\ \hline
\multicolumn{2}{|c|}{Reliability-related Open Research Questions}                                         \\ \hline
How to mitigate camera-LiDAR coupling?                & Sensor-agnostic framework                         \\ \hline
How to improve all-weather / Lighting conditions?     & Datasets with complex weather / lighting conditions \\ \hline
How to handle adversarial attacks and corner cases?   & Cross modality verification                       \\ \hline
How to solve open-set object detection?               & Testing protocol, metrics + new framework         \\ \hline
How to balance speed-accuracy trade-offs?             & Models developed with scalability in mind\\ \hline
\end{tabular}
\end{table*}

\section{Trends, Open Challenges and Promising Directions}

\par The perception module in a driverless car is responsible for obtaining and understanding its surrounding scenes. Its down-stream modules, such as planning, decision making and self-localization, depend on its outputs. Therefore, its performance and reliability are the prerequisites for the competence of the entire driverless system. To this end, LiDAR and camera fusion is applied to improve the performance and reliability of the perception system, making driverless vehicles more capable in understanding complex scenes (e.g. urban traffic, extreme weather condition and so on). Consequently, in this section, we summarize overall trends and discuss open challenges and potential influencing factors in this regard. As shown in Table VI, we focus on improving the performance of fusion methodology and the robustness of the fusion pipeline.

\par From the methods reviewed above, we observed some general trends among the image and point cloud fusion approaches, which are summarized as the following: 
\begin{itemize}
\item 2D to 3D: Under the progressing of 3D feature extraction methods, to locate, track and segment objects in 3D space has become a heated area of research.
\item Single-task to multi-tasks: Some recent works \cite{Liang_2019_CVPR} \cite{simon2019complexeryolo} combined multiple complementary tasks, such as object detection, semantic segmentation and depth completion to achieve better overall performance and reduce computational costs.  
\item Signal-level to multi-level fusion: Early works often leverage signal-level fusion where 3D geometry is translated to the image plane to leverage off-the-shelf image processing models, while recent models try to fuse image and LiDAR in multi-level (e.g. early fusion, late fusion) and temporal context encoding.
\end{itemize}

\subsection{Performance-related Open Research Questions}
\subsubsection{What should be the data representation of fused data?}

\par The choosing of data representations of the fused data plays a fundamental role in designing any data fusion algorithms. 
Current data representation for image and point cloud fusion includes: 
\begin{itemize}
\item \textbf{Image Representation:} Append 3D geometry as additional channels of the image. The image-based representation enables off-the-shelf image processing models. However, the results are also limited in the 2D image plane, which is less ideal for autonomous driving. 
\item \textbf{Point Representation:} Append RGB signal/features as additional channels of the point cloud. However, the mismatch of resolution between high-resolution images and low-resolution point cloud leads to inefficiency. 
\item \textbf{Intermediate data representations:} Translate image and point cloud features/signal into an intermediate data representation, such as voxelized point cloud \cite{10.1007/978-3-319-10599-4_41}. However, voxel-based methods suffer from bad scalability.
\end{itemize}
\par Many recent works for point cloud processing have concentrated on defining explicit point convolution operations \cite{hua2018pointwise} \cite{Xu_2018} \cite{Atzmon_2018} \cite{NIPS2018_7362} \cite{Lan_2019} \cite{Groh_2019} \cite{Thomas_2019}, which have shown great potentials. These point convolutions are better suited to extract fine-grained per-point and local geometry. Therefore, we believe the point representation of fused data coupled with point convolutions has great potentials in camera-LiDAR fusion studies.

% \subsubsection{How to encode geometric constraint?}
% \par Compared with other sources of depth data, such as RGBD data from stereo or structured light, LiDAR has longer rangeability and higher accuracy, which provide detailed and accurate 3D geometry. The geometric constraint has become common sense in the fusion pipeline of image and point cloud, which provide extra information to guide the network to achieve better performance.
% \par Projecting point cloud to the image plane in the form of RGBD image seems the most natural workaround for point cloud's unordered data format, but the sparsity attribute would cause empty holes. Depth completion and point cloud up-sampling could handle this problem to some extent. On the other hand, methods in monocular depth prediction introduce self-supervised learning between consecutive frames that may hopefully ease the situation. However, how to encoding this geometry into the fusion pipeline remains to be explored.

\subsubsection{How to encode temporal context?}

\par Most current deep learning based perception systems tend to overlook temporal context. This leads to numerous problems, such as point cloud deformation from low refresh rate and incorrect time-synchronization between sensors. These problems cause mismatches between images, point cloud and actual environment. Therefore, it is vital to incorporate temporal context into the perception systems.

\par In the context of autonomous driving, temporal context can be incorporated using RNN or LSTM models. In \cite{8500658}, an LSTM auto-encoder is employed to estimate future states of surrounding vehicles and adjust the planned trajectory accordingly, which helps autonomous vehicles run smoother and more stable. In \cite{Luiten_2020} temporal context is leveraged to estimate ego-motion, which benefits later task-related header networks. Furthermore, the temporal context could benefit online self-calibration via a visual-odometry based method \cite{chien2016visual}. Following this trend, the mismatches caused by LiDAR low refresh rate could be solved by encoding temporal context and generative models.

\subsubsection{What should be the learning scheme?}
\par Most current camera-LiDAR fusion methods rely on supervised learning, which requires large annotated datasets. However, annotating images and point clouds is expensive and time-consuming. This limits the size of the current multi-modal dataset and the performance of supervised learning methods. 
\par The answer to this problem is unsupervised and weakly-supervised learning frameworks. Some recent studies have shown great potentials in this regards \cite{Ma_2019} \cite{Cheng_2019} \cite{287150} \cite{8931245} \cite{7378972}. Following this trend, future researches in unsupervised and weakly-supervised learning fusion frameworks could allow the networks to be trained on large unlabeled/coarse-labelled dataset and leads to better performance.

\subsubsection{When to use deep learning methods?}
\par Recent advances in deep learning techniques have accelerated the development of autonomous driving technology. In many aspects, however, traditional methods are still indispensable in current autonomous driving systems. Compared with deep learning methods, traditional methods have better interpretability and consume significantly less computational resources. The ability to track back a decision is crucial for the decision making and planning system of an autonomous vehicle. Nevertheless, current deep learning algorithms are not back-traceable, making them unfit for these applications. Apart from this black-box dilemma, traditional algorithms are also preferred for their real-time capability.
\par To summarize, we believe deep learning methods should be applied to applications that have explicit objectives that can be verified objectively.

\subsection{Reliability-related Open Research Questions}
\subsubsection{How to mitigate camera-LiDAR coupling?}
\par From an engineering perspective, redundancy design in an autonomous vehicle is crucial for its safety.  Although fusing LiDAR and camera improves perception performance, it also comes with the problem of signal coupling. If one of the signal paths suddenly failed, the whole pipeline could break down and cripple down-stream modules. This is unacceptable for autonomous driving systems, which require robust perception pipelines.

\par To solve this problem, we should develop a sensor-agnostic framework. For instance, we can adopt multiple fusion modules with different sensor inputs. Furthermore, we can employ a multi-path fusion module that take asynchronous multi-modal data. However, the best solution is still open for study.

\subsubsection{How to improve all-weather/Lighting conditions?}

\par Autonomous vehicles need to work in all weather and lighting conditions. However, current datasets and methods are mostly focused on scenes with good lighting and weather conditions. This leads to bad performances in the real-world, where illumination and weather conditions are more complex.
\par The first step towards this problem is developing more datasets that contain a wide range of lighting and weather conditions. In addition, methods that employ multi-modal data to tackle complex lighting and weather conditions require further investigation. 

\subsubsection{How to handle adversarial attacks and corner cases?}

\par Adversarial attacks targeted at camera-based perception systems have proven to be effective. This poses a grave danger for the autonomous vehicle, as it operates in safety-critical environments. It may be difficult to identify attacks explicitly designed for certain sensory modality. However, perception results can be verified across different modalities. In this context, research on utilizing the 3D geometry and images to jointly identify these attacks can be further explored. 
\par As self-driving cars operate in an unpredictable open environment with infinite possibilities, it is critical to consider corner and edge cases in the design of the perception pipeline. The perception system should anticipate unseen and unusual obstacles, strange behaviours and extreme weather. For instance, the image of cyclists printed on a large vehicle and people wearing costumes. These corner cases are often very difficult to handle using only the camera or LiDAR pipeline. However, leveraging data from multi-modalities to identify these corner cases could prove to be more effective and reliable than from a-modal sensors. Further researches in this direction could greatly benefit the safety and commercialization of autonomous driving technology.

\subsubsection{How to solve open-set object detection?}

\par Open-set object detection is a scenario where an object detector is tested on instances from unknown/unseen classes. The open-set problem is critical for an autonomous vehicle, as it operates in unconstrained environments with infinite categories of objects. Current datasets often use a background class for any objects that are not interested. However, no dataset can include all the unwanted object categories in a background class. Therefore, the behaviour of an object detector in an open-set setting is highly uncertain, which is less ideal for autonomous driving. 

\par The lack of open-set object detection awareness, testing protocol and metrics leads to little explicit evaluations of the open-set performance in current object detection studies. These challenges have been discussed and investigated in a recent study by Dhamija et al. \cite{9093355}, where a novel open-set protocol and metrics are proposed. The authors proposed an additional mixed unknown category, which contains known 'background' objects and unknown/unseen objects. Based on this protocol, current methods are tested on a test set with a mixed unknown category generated from a combination of existing datasets. In another recent study on the point cloud, Wong et al. \cite{wong2020identifying} proposed a technique that maps unwanted objects from different categories into a category-agnostic embedding space for clustering.
\par The open-set challenges are essential for deploying deep learning-based perception systems in the real-world. And it needs more effort and attention from the whole research community (dataset and methods with emphasis on unknown objects, test protocols and metrics, etc.,).

\subsubsection{How to balance speed-accuracy trade-offs?}
\par The processing of multiple high-resolution images and large-scale point clouds put substantial stress on existing mobile computing platforms. This sometimes causes frame-drop, which could seriously degrade the performance of the perception system. More generally, it leads to high-power consumption and low reliability. Therefore, it is important to balance the speed and accuracy of a model in real-world deployments.
\par There are studies that attempt to detect the frame-drop. In \cite{6375020}, Imre et al. proposed a multi-camera frame-drop detection algorithm that leverages multiple segments (broken line) fitting on camera pairs. However, frame-drop detection only solves half of the problem. The hard-part is to prevent the performance degradation caused by the frame-drop. Recent advances in generative models have demonstrated great potentials to predict missing frames in video sequences \cite{li2019here}, which could be leveraged in autonomous driving for filling the missing frames in the image and the point cloud pipelines. However, we believe the most effective way to solve the frame-drop problem is to prevent it by reducing the hard-ware workload. This can be achieved by carefully balancing the speed and accuracy of a model \cite{Huang_2017}.
\par To achieve this, deep learning models should be able to scale down its computational cost, while maintaining acceptable performance. This scalability is often achieved by reducing the number of inputs (points, pixels, voxels) or the depth of the network. From previous studies \cite{qi2017pointnet} \cite{Zhirong_Wu_2015} \cite{Thomas_2019}, point-based and multi-view based fusion methods are more scalable compared to voxel-based methods.
% \par Nevertheless, current studies on camera-LiDAR fusion often neglect the importance of model scalability. This is because methods often aim to out-perform the SOTA, while real-world deployment receives little attention. We believe incorporating standard tests on the model scalability would facilitate the real-world deployment of deep learning models.

\section{Conclusion}
\par This paper presented an in-depth review of the most recent progress on deep learning models for point cloud and image fusion in the context of autonomous driving. Specifically, this review organizes methods based on their fusion methodologies and covers topics in depth completion, dynamic and stationary object detection, semantic segmentation, tracking and online cross-sensor calibration. Furthermore, performance comparisons on the publicly available dataset, highlights and advantages/disadvantages of models are presented in tables. Typical model architectures are shown in figures. Finally, we summarized general trends and discussed open challenges and possible future directions. This survey also raised awareness and provided insights on questions that are overlooked by the research community but troubles real-world deployment of the autonomous driving technology.

\bibliographystyle{IEEEtran}
\bibliography{IEEEabrv,Bibliography}

%\end{thebibliography}
%\begin{biography}[{\includegraphics[width=1in,height=1.25in,clip,keepaspectratio]{mshell}}]{Michael Shell}

% ==== SWITCH OFF the BIO for submission
% ==== SWITCH OFF the BIO for submission

%\begin{IEEEbiographynophoto}{Ignacio Ramos}
%(S'12) received the B.S. degree in electrical engineering from the University of Illinois at Chicago in 2009, and is currently working toward the Ph.D. degree at the University of Colorado at Boulder. From 2009 to 2011, he was with the Power and Electronic Systems Department at Raytheon IDS, Sudbury, MA. His research interests include high-efficiency microwave power amplifiers, microwave DC/DC converters, radar systems, and wireless power transmission.

%% insert where needed to balance the two columns on the last page with
%%\newpage

% ==== SWITCH OFF the BIO for submission
% ==== SWITCH OFF the BIO for submission

\begin{IEEEbiography}[{\includegraphics[width=1in,height=1.25in,clip,keepaspectratio]{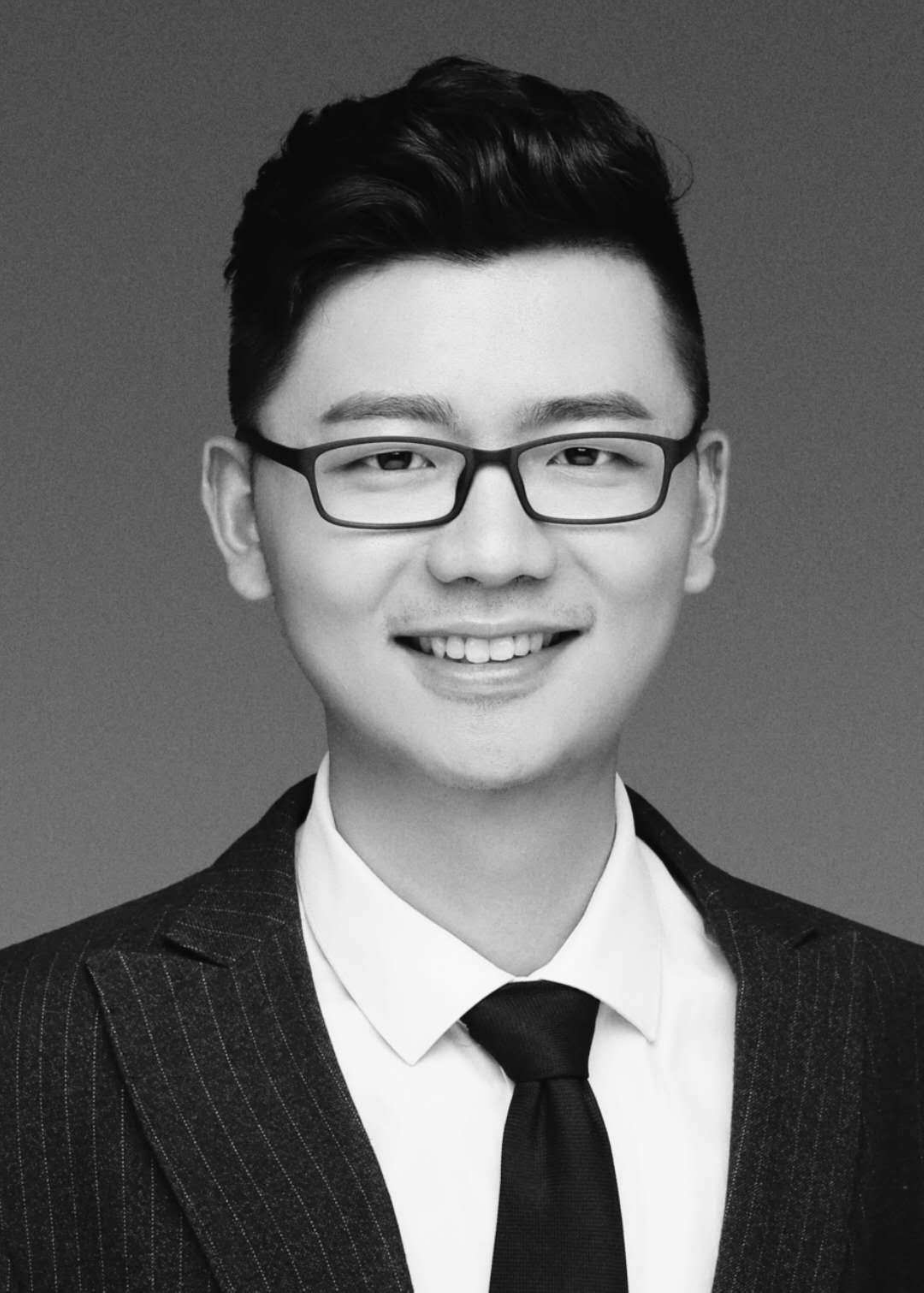}}]{Yaodong Cui}
received the B.S. degree in automation from Chang'an University, Xi'an, China, in 2017, received the M.Sc. degree in  Systems, Control and Signal Processing from the University of Southampton, Southampton, UK, in 2019. He is currently working toward a Ph.D. degree with the Waterloo Cognitive Autonomous Driving (CogDrive) Lab, Department of Mechanical Engineering, University of Waterloo, Waterloo, Canada. His research interests include sensor fusion, perception for the intelligent vehicle, driver emotion detection.
\end{IEEEbiography}

\begin{IEEEbiography}[{\includegraphics[width=1in,height=1.25in,clip,keepaspectratio]{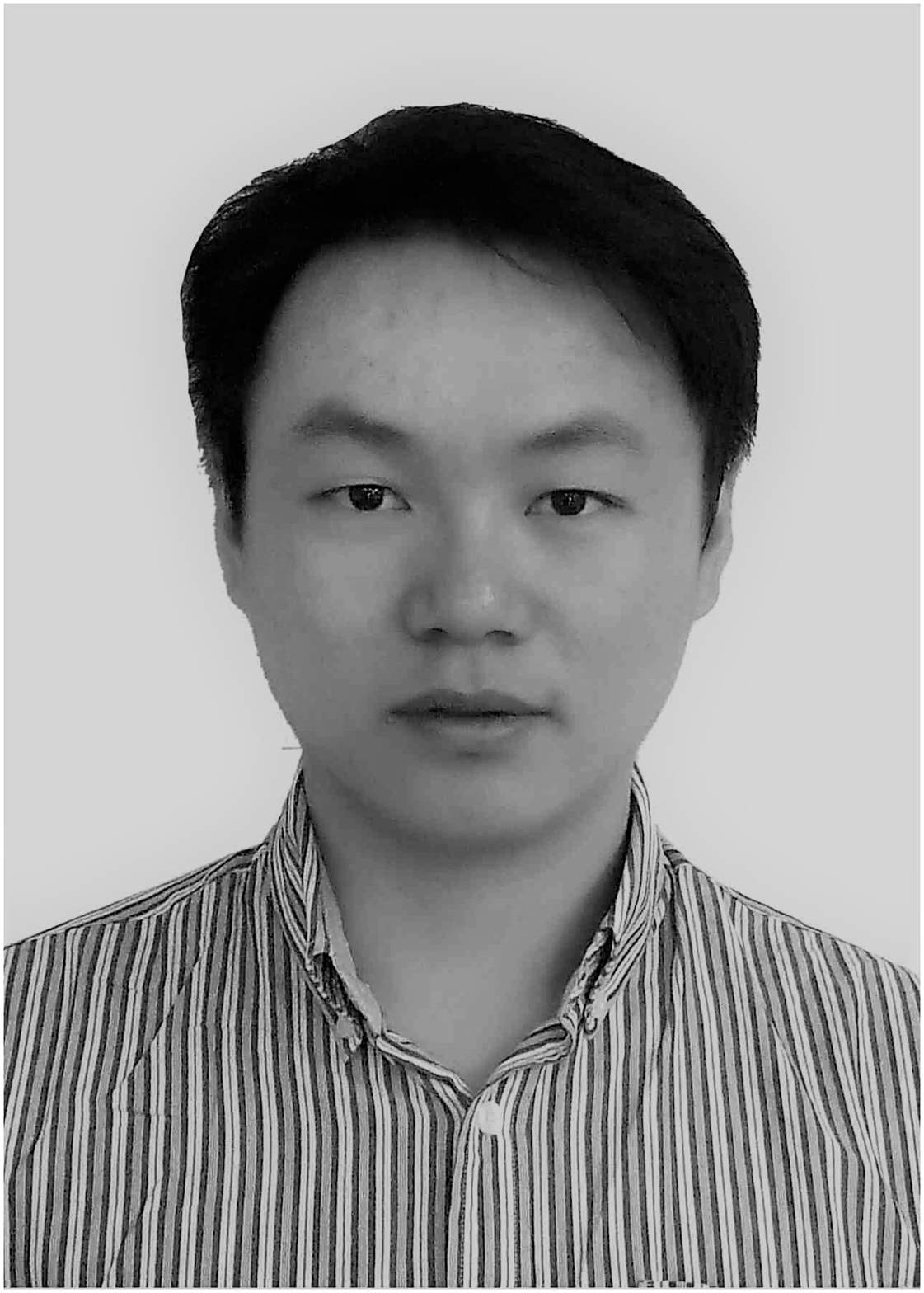}}]{Ren Chen}
received the B.S. degree in electronic science and technology from Huazhong University of Science and Technology, Wuhan, China. received the M.Sc. degree in  Computer Science from China Academy of Aerospace Science and Engineering, Beijing, China. He worked at Institude of Deep Learning, Baidu and Autonomous Driving Lab, Tencent. His research interests include computer vision, perception for robotics, realtime dense 3D reconstruction and robotic motion planning.
\end{IEEEbiography}

\begin{IEEEbiography}[{\includegraphics[width=1in,height=1.25in,clip,keepaspectratio]{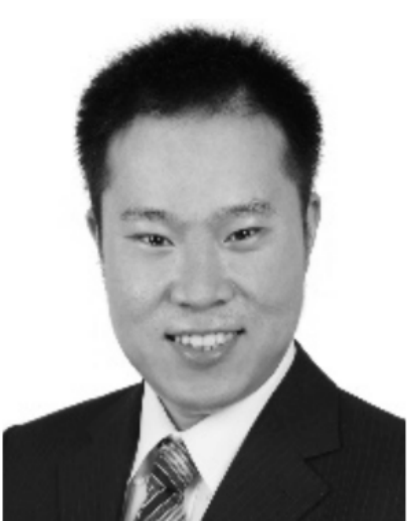}}]{Wenbo Chu}
received the B.S. degree in automotive engineering from Tsinghua University, Beijing, China, in 2008, the M.S. degree in automotive engineering from RWTH Aachen, Germany, and the Ph.D. degree in mechanical engineering from Tsinghua University, in 2014. He is currently a Research Fellow of China Intelligent and Connected Vehicles Research Institute Company, Ltd., Beijing, which is also known as the National Innovation Center of Intelligent and
Connected Vehicles. His research interests include intelligent connected vehicles, new energy vehicles, and vehicle dynamics and control
\end{IEEEbiography}

\begin{IEEEbiography}[{\includegraphics[width=1in,height=1.25in,clip,keepaspectratio]{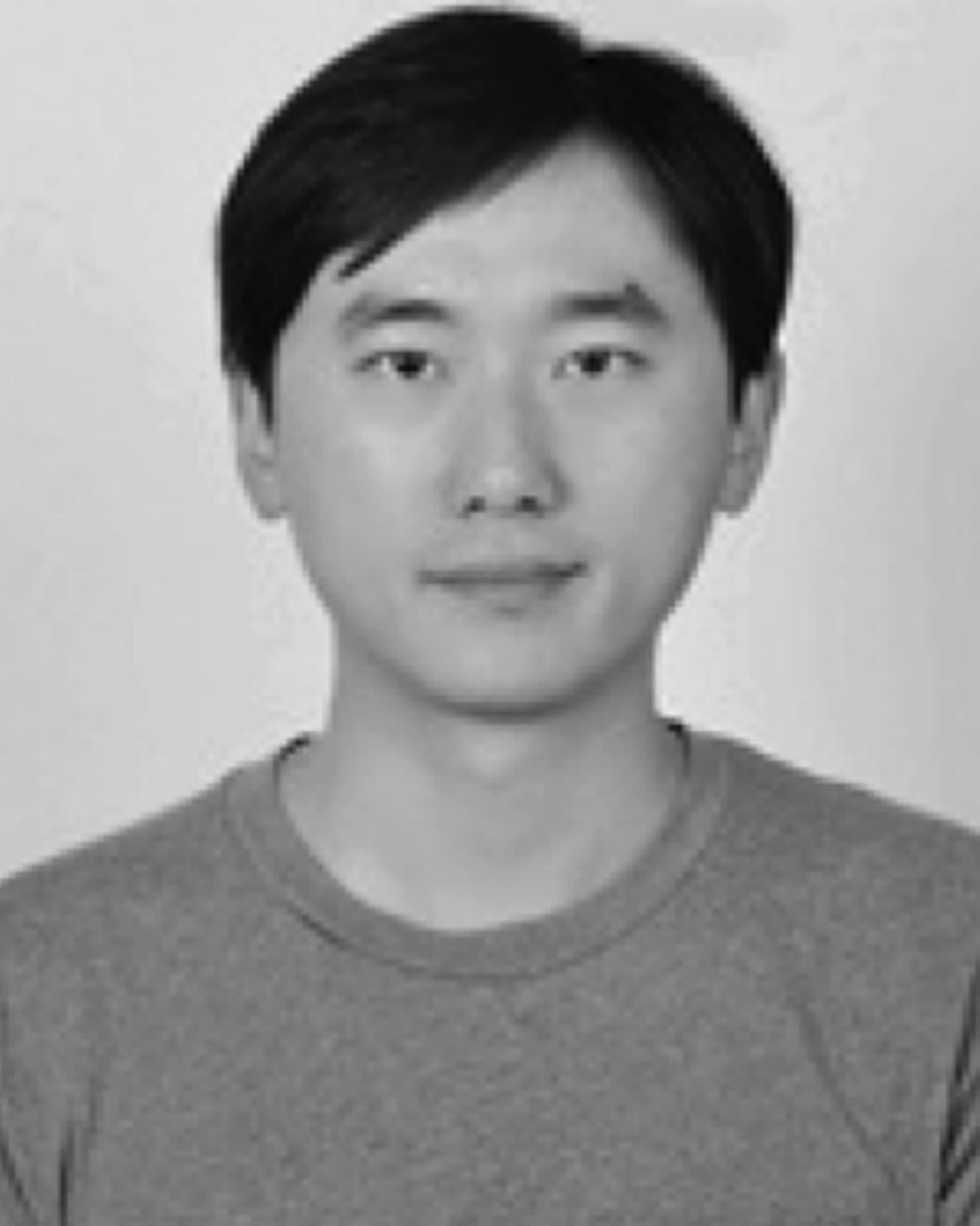}}]{Long Chen}
received the B.Sc. degree in communication engineering and the Ph.D. degree in signal and information processing from Wuhan University, Wuhan, China.He is currently an Associate Professor with School of Data and Computer Science, Sun Yat-sen University, Guangzhou, China.
%He received the IEEE Vehicular Technology Society 2018 Best Land Transportation Paper Award.
His areas of interest include autonomous driving, robotics, artificial intelligence where he has contributed more than 70 publications.
He serves as an Associate Editor for IEEE Transactions on Intelligent Transportation Systems.
\end{IEEEbiography}

\begin{IEEEbiography}[{\includegraphics[width=1in,height=1.25in,clip,keepaspectratio]{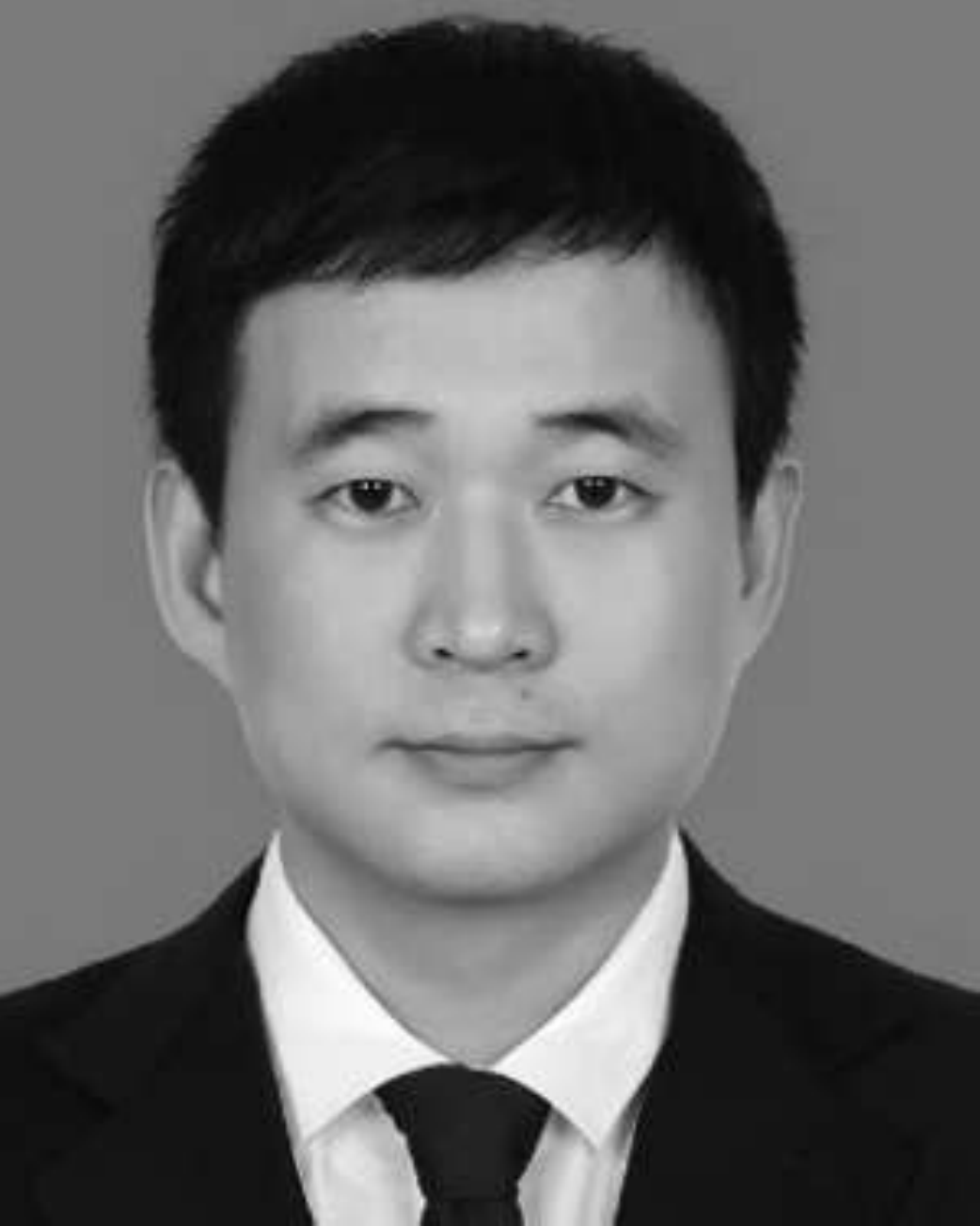}}]{Daxin Tian}
is currently a professor in the School of Transportation Science and Engineering, Beihang University, Beijing, China. He is IEEE Senior Member, IEEE Intelligent Transportation Systems Society Member, and IEEE Vehicular Technology Society Member, etc. His current research interests include mobile computing, intelligent transportation systems, vehicular ad hoc networks, and swarm intelligent.
\end{IEEEbiography}

\begin{IEEEbiography}[{\includegraphics[width=1in,height=1.25in,clip,keepaspectratio]{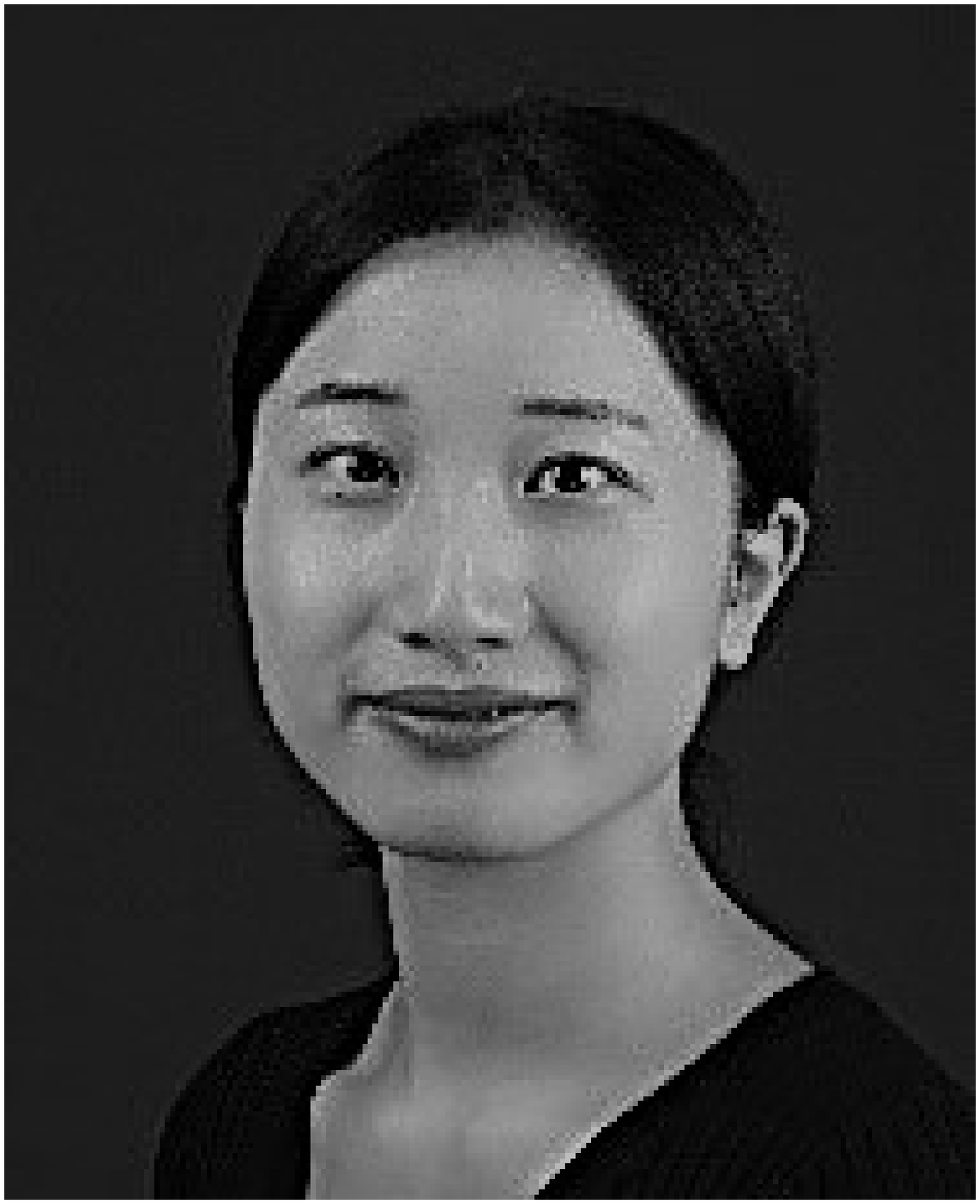}}]{Ying Li}
received the M.Sc. degree in remote sensing from Wuhan University, China, in 2017. She is currently pursuing the Ph.D. degree with the Mobile Sensing and Geodata Science Laboratory, Department of Geography and Environmental Management, University of Waterloo, Canada.
Her research interests include autonomous driving, mobile laser scanning, intelligent processing of point clouds, geometric and semantic modeling, and augmented reality.
\end{IEEEbiography}

\begin{IEEEbiography}[{\includegraphics[width=1in,height=1.25in,clip,keepaspectratio]{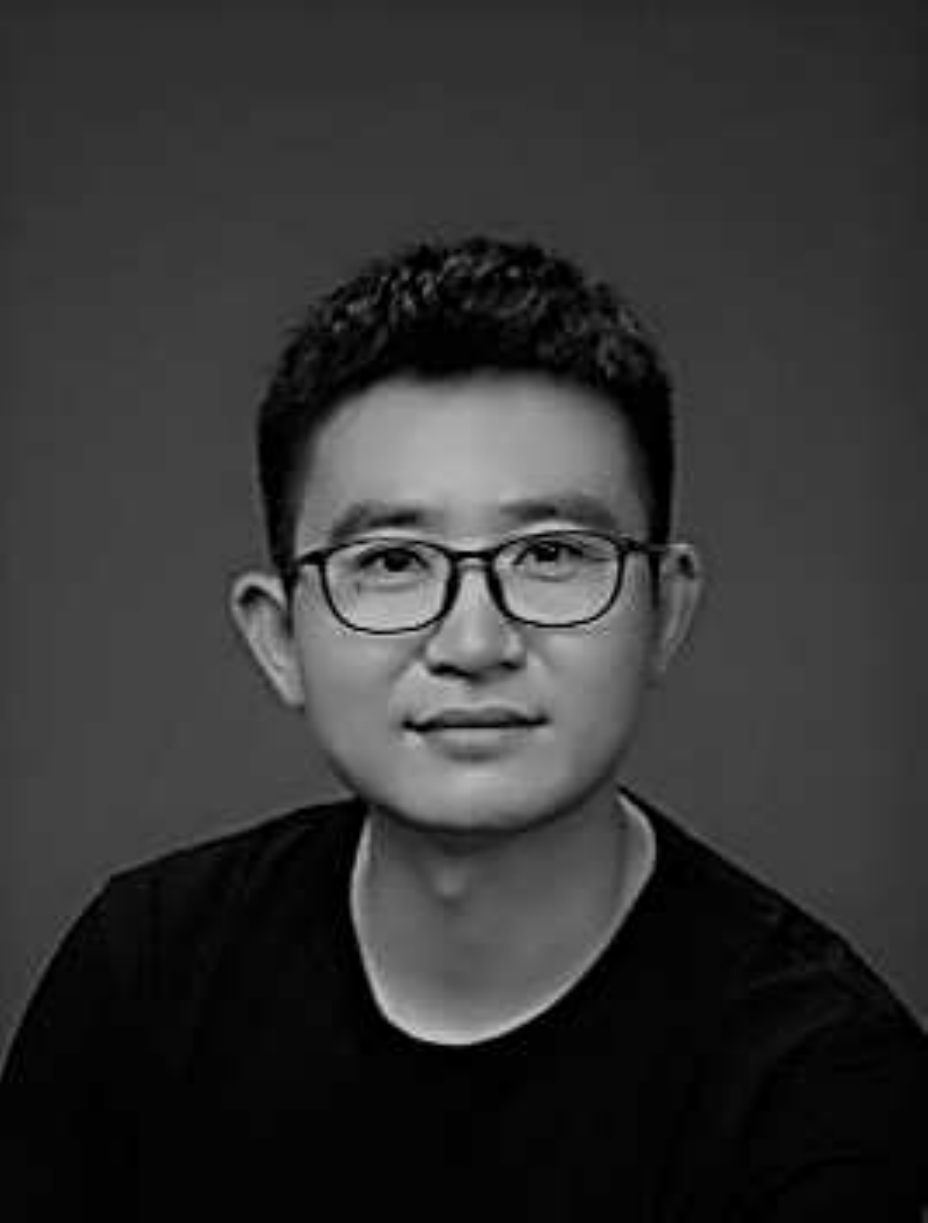}}]{Dongpu Cao}
received the Ph.D. degree from Concordia University, Canada, in 2008. He is the Canada Research Chair in Driver Cognition and Automated Driving, and currently an Associate Professor and Director of Waterloo Cognitive Autonomous Driving (CogDrive) Lab at University of Waterloo, Canada. His current research focuses on driver cognition, automated driving and cognitive autonomous driving. He has contributed more than 200 papers and 3 books. He received the SAE Arch T. Colwell Merit Award in 2012, IEEE VTS 2020 Best Vehicular Electronics Paper Award and three Best Paper Awards from the ASME and IEEE conferences. Prof. Cao serves as Deputy Editor-in-Chief for IET INTELLIGENT TRANSPORT SYSTEMS JOURNAL, and an Associate Editor for IEEE TRANSACTIONS ON VEHICULAR TECHNOLOGY, IEEE TRANSACTIONS ON INTELLIGENT TRANSPORTATION SYSTEMS, IEEE/ASME TRANSACTIONS ON MECHATRONICS, IEEE TRANSACTIONS ON INDUSTRIAL ELECTRONICS, IEEE/CAA JOURNAL OF AUTOMATICA SINICA, IEEE TRANSACTIONS ON COMPUTATIONAL SOCIAL SYSTEMS, and ASME JOURNAL OF DYNAMIC SYSTEMS, MEASUREMENT AND CONTROL. He was a Guest Editor for VEHICLE SYSTEM DYNAMICS, IEEE TRANSACTIONS ON SMC: SYSTEMS and IEEE INTERNET OF THINGS JOURNAL. He serves on the SAE Vehicle Dynamics Standards Committee and acts as the Co-Chair of IEEE ITSS Technical Committee on Cooperative Driving. Prof. Cao is an IEEE VTS Distinguished Lecturer
\end{IEEEbiography}


\vfill


\end{document}